\documentclass[lettersize,journal,twoside]{IEEEtran}  %

\IEEEoverridecommandlockouts  %

\usepackage{amscd}
\usepackage{amssymb}
\usepackage{epsf}
\usepackage{pgf}
\usepackage{url}

\usepackage{amsmath,amsfonts}
\usepackage[T1]{fontenc}  %
\usepackage{psfrag}
\usepackage{algorithm}  %
\usepackage{algpseudocode}
\usepackage{pifont}
\usepackage{cite}
\usepackage[hidelinks]{hyperref}  %
\usepackage{booktabs}
\usepackage[amsmath,hyperref,thmmarks]{ntheorem}  %
\usepackage{pgfgantt}
\usepackage{import}
\usepackage{listings} 
\usepackage{bm} %
\usepackage{bbm} %
\usepackage{footnote} %
\usepackage{dirtree} %
\usepackage{seqsplit} %
\usepackage{microtype}  %
\usepackage{placeins}  %

\usepackage{tikz}
\usetikzlibrary{ext.paths.ortho}  %
\usetikzlibrary{backgrounds,fit}
\usetikzlibrary {decorations.pathreplacing} 
\usepackage{pgfplots}
\pgfplotsset{compat=newest}
\usepgfplotslibrary{groupplots}
\usepgfplotslibrary{fillbetween}
\usetikzlibrary{shapes.misc}

\usepackage{tumcolor}
\usepackage{tabularx}
\usepackage{multirow}
\usepackage{array}   %
\usepackage{xstring}
\newcolumntype{L}{>{$}l<{$}} %
\newcolumntype{C}{>{$}c<{$}} %
\newcolumntype{R}{>{$}r<{$}} %
\usepackage{pifont}  %
\newcommand{\cmark}{\ding{51}}%
\newcommand{\xmark}{\ding{55}}%
\usepackage{xfrac}
\usepackage{enumitem}  %
\setlist{parsep=0pt,listparindent=\parindent}
\newlist{hypothesis}{enumerate}{1}  %
\setlist[hypothesis,1]{
    label=\textbf{H\arabic*},  %
    ref=\textbf{H\arabic*},  %
    resume}  %
\newlist{lexicographic}{enumerate}{1}
\setlist[lexicographic,1]{
    label=\arabic*),  %
    ref=\arabic*}  %

\usepackage{romannum}  %

\usepackage[separate-uncertainty = true]{siunitx}  %
\usepackage[nameinlink,capitalise]{cleveref}  %
\crefname{algorithm}{Alg.}{Algs.}
\crefname{figure}{Fig.}{Figs.}
\crefname{section}{Sec.}{Sec.}
\crefname{table}{Tab.}{Tab.}
\crefname{equation}{}{}  %
\crefname{appendix}{App.}{Apps.}
\crefformat{footnote}{\textsuperscript{#2#1#3}}

\theorembodyfont{\normalfont}
\theoremstyle{plain}
\theoremsymbol{\ensuremath{\square}}
\theoremheaderfont{\bf}
\theoremseparator{:}

\DeclareMathOperator*{\argmin}{arg\,min}

\newcommand{\norm}[1]{\left\lVert#1\right\rVert}
\newcommand{\normElem}[1]{\left|#1\right|}

\newcommand{\indicator}{\mathbb{I}}
\newcommand{\indicatorF}[1]{\indicator\left( #1 \right)}

\newcommand{\eef}{\mathrm{eef}}
\newcommand{\desired}{\mathrm{d}}
\newcommand{\external}{\mathrm{ext}}
\newcommand{\dynamics}{\mathrm{dyn}}

\newcommand{\duration}{\mathrm{d}}
  
\newcommand{\occ}{\mathrm{occ}}

\DeclareMathOperator{\otherwise}{otherwise}

\newcommand{\bools}{\mathbb{B}}  
\newcommand{\reals}{\mathbb{R}}  
\newcommand{\naturals}{\mathbb{N}}  
\newcommand{\constraints}{\mathcal{C}}  
\newcommand{\G}{\mathcal{G}}  
\newcommand{\workspace}{\mathcal{W}}  
\newcommand{\obstacle}{\workspace_{\occ}}
\newcommand{\occupancy}{\mathcal{O}}  
\newcommand{\powerset}{\mathcal{P}}  

\newcommand{\SEThree}{\mathbb{SE}(3)}
\newcommand{\SETwo}{\mathbb{SE}(2)}
\newcommand{\SOThree}{\mathbb{SO}(3)}
\newcommand{\SOTwo}{\mathbb{SO}(2)}
\newcommand{\IKsols}{\mathcal{Q}}
\newcommand{\validBases}{\mathcal{B}}
\newcommand{\naturalInt}[1]{\left< #1 \right>}
\newcommand{\realInt}[2]{\left< #1, #2 \right>}
\newcommand{\normalDist}[2]{\mathcal{N}_{#1, #2}}

\newcommand{\modules}{\mathcal{R}}
\newcommand{\costid}{\mathrm{C}}
\newcommand{\costfun}[1]{J_{#1}}
\newcommand{\goal}[1][]{
    \ifthenelse{\equal{#1}{}}{g}{g_{#1}}
}
\newcommand{\constraint}[1][]{
    \ifthenelse{\equal{#1}{}}{c}{c_{#1}}
}

\newcommand{\T}{\mathbf{T}}  
\newcommand{\Teff}{\T_{\eef}}
\newcommand{\Tdes}[1][]{
    \ifthenelse{
        \equal{#1}{}
    }{\T_{\desired}
    }{\T_{\desired, #1}}
}
\newcommand{\B}{\mathbf{B}}  

\newcommand{\Proj}{\mathbf{s}}
  
\newcommand{\rotM}{\mathbf{R}}

\newcommand{\Jposv}{\mathbf{q}}  
\newcommand{\Jposvmax}[1][]{
    \ifthenelse{\equal{#1}{}}
    {\Jposv_\mathrm{max}}
    {\Jposv_{\mathrm{max}, #1}}}
\newcommand{\Jposvmin}[1][]{
    \ifthenelse{\equal{#1}{}}
    {\Jposv_\mathrm{min}}
    {\Jposv_{\mathrm{min}, #1}}}
\newcommand{\Jvelv}{\dot{\mathbf{q}}}  
\newcommand{\Jvelvmax}[1][]{
    \ifthenelse{\equal{#1}{}}
    {\Jvelv_\mathrm{max}}
    {\Jvelv_{\mathrm{max}, #1}}
}
\newcommand{\Jaccv}{\ddot{\mathbf{q}}}  
\newcommand{\Jaccvmax}{\Jaccv_\mathrm{max}}
\newcommand{\Jtorqv}{\bm{\tau}}  
\newcommand{\Jtorqvmax}{\Jtorqv_\mathrm{max}}  
\newcommand{\Jmotionv}[1][]{
    \ifthenelse{\equal{#1}{}}{\mathbf{z}_{\desired}}{\mathbf{z}_{\desired,#1}}
}
\newcommand{\Jpos}{q}

\newcommand{\Jposs}{\mathbf{Q}}

\newcommand{\Transv}{\mathbf{t}}
\newcommand{\Pointv}{\mathbf{p}}
\newcommand{\qpath}{\hat{\Jposv}}
\newcommand{\Fext}{\mathbf{f}_{\external}}
\newcommand{\Zerov}{\mathbf{0}}

\newcommand{\Tolbound}{\gamma}

\newcommand{\basev}{\mathbf{b}}
\newcommand{\linesegv}{\mathbf{l}}

\newcommand{\timenow}{t}
\newcommand{\timezero}{0}
\newcommand{\timeend}[1][]{
    \ifthenelse{\equal{#1}{}}{t_{\mathrm{N}}}{t_{\mathrm{N}, #1}}
}
\newcommand{\timeany}{t'}  

\newcommand{\timeduration}[1][]{
    \ifthenelse{\equal{#1}{}}{t_{\duration}}{t_{\duration}( #1 )}
}  
\newcommand{\ttimeout}{t_{\mathrm{CPU}}}

\newcommand{\link}{l}
\newcommand{\assembly}{\mathbf{m}}
\newcommand{\module}[1]{m_{#1}}

\newcommand{\dof}{n_{\mathrm{DoF}}}
\newcommand{\nModules}{n_{\mathrm{mod}}}

\newcommand{\nSimpleIK}{n_{\mathrm{IK}}}
\newcommand{\nCplxIK}{n_{\mathrm{IK, Obs}}}
\newcommand{\tplan}{t_{\mathrm{plan}}}

\newcommand{\sigmaBase}{\sigma_{\mathrm{\B}}}
\newcommand{\sigmaIK}{\sigma_{\mathrm{IK}}}
\newcommand{\pLamarck}{p_{\mathrm{Lamarck}}}
\newcommand{\nLamarck}{n_{\mathrm{Lamarck}}}

\newcommand{\dofInit}{n_{\mathrm{DoF, init}}}
\newcommand{\chanceEmpty}{p_{\mathrm{empty}}}
\newcommand{\popSize}{n_{\mathrm{pop}}}
\newcommand{\nGenes}{n_{\mathrm{genes}}}
\newcommand{\pMutate}{p_{\mathrm{mutate}}}
\newcommand{\nParentsMate}{n_{\mathrm{parents, mate}}}
\newcommand{\nKeepParents}{n_{\mathrm{parents, keep}}}
\newcommand{\nElites}{n_{\mathrm{elites}}}
\newcommand{\selectionPres}{p_{\mathrm{select}}}

\newcommand{\nBase}{{n_{\mathrm{\B}}}}

\newcommand{\timeJ}{J_{\mathrm{T}}}  
\newcommand{\costfail}{J_{\mathrm{fail}}}

\newcommand{\scopeM}{\assembly}
\newcommand{\scopeMB}{\assembly + \B}
\newcommand{\scopeMQ}{\assembly + \Jposs}
\newcommand{\scopeMBQ}{\assembly + \B + \Jposs}

\DeclareSIUnit[number-unit-product = {}]{\inch}{''}

\newcommand{\lastChecked}{Accessed: Oct. 28$^{\text{th}}$, 2025}

\usepackage{censor}

\usepackage{graphicx}
\graphicspath{ {./images/} }

\usepackage{tikz}
\newcommand{\drawRobot}[4]{
    
    \typeout{Drawing robot with base at #1}
    \IfBeginWith{#1}{(}{
        \typeout{Base given relative: #1}
        \coordinate (P0) at #1;
        \pgfextractx{\xzero}{\pgfpointanchor{#1}{center}}
        \pgfextracty{\yzero}{\pgfpointanchor{#1}{center}}
    }{
        \pgfmathsetlengthmacro{\xzero}{#1[0]}
        \pgfmathsetlengthmacro{\yzero}{#1[1]}
        \coordinate (P0) at (\xzero,\yzero);
    }
    \typeout{Base position: (\xzero, \yzero)}
    \pgfmathsetlengthmacro{\radiusJoint}{#4}
    \typeout{Joint radius position: \radiusJoint}
    \pgfmathsetmacro{\linkNum}{int(dim(#2) - 1)}
    \typeout{Will draw link 0 to \linkNum}

    \def\lengths{#2}
    \def\angles{#3}
    \typeout{Lengths \lengths}
    \typeout{Angles \angles}
    
    \filldraw[black] (P0) circle (\radiusJoint); %
    
    \foreach \i [
        remember=\x as \lastx (initially \xzero), 
        remember=\y as \lasty (initially \yzero),
        remember=\t as \lastt (initially 0)] in {0,...,\linkNum} {
        \typeout{Iter \i, x=\lastx, y=\lasty, t=\lastt}
        \pgfmathsetlengthmacro{\length}{\lengths[\i]}
        \pgfmathsetmacro{\angle}{\angles[\i]}
        \typeout{Drawing link of length \length angle \angle}
        
        \pgfmathsetmacro{\t}{\lastt + \angle}
        \typeout{Global angle \t}
        \pgfmathsetlengthmacro{\x}{\lastx + (\length * cos(\t))}
        \pgfmathsetlengthmacro{\y}{\lasty + (\length * sin(\t))}
        \pgfmathsetmacro{\nexti}{int(\i + 1)}
        \typeout{Next point at x = \x}
        \typeout{Next point at y = \y}
        \typeout{Draw from P\i to node P\nexti}
        \coordinate (P\nexti) at (\x,\y);
        
        \draw[-] (P\i) -- (P\nexti);
        
        \filldraw[black] (P\nexti) circle (\radiusJoint); %
    }
}

\newcommand{\drawEEF}[3]{
    \path #1 -- ++(#2:2*#3) -- ([turn]-0.5*#3,0.5*#3) coordinate (one) -- ([turn]0.7071*#3,0*#3) coordinate (two) -- ([turn]0*#3,1.414*#3) coordinate (three) -- ([turn]0*#3,1.414*#3) coordinate (four) -- ([turn]0*#3,0.7071*#3) coordinate (five);
    \draw (one) -- (two) -- (three) -- (four) -- (five);
}

\StopCensoring  %

\newcommand\copyrighttext{%
  \footnotesize \textcopyright 2025 The Authors. This work is licensed under a Creative Commons Attribution 4.0 License.\\[0pt]
  \footnotesize For more information, see \href{https://creativecommons.org/licenses/by/4.0}{creativecommons.org/licenses/by/4.0/}.}

\title{\LARGE \bf
Holistic Optimization of Modular Robots}

\author{
    Matthias Mayer and Matthias Althoff, \IEEEmembership{Member, IEEE}%
    \thanks{
        Manuscript received April 30$^\text{th}$, 2025; revised Aug. 15$^\text{th}$, 2025; accepted Oct. 1$^\text{st}$, 2025.
        Date of publication 5 November 2025; date of current version 2 February 2026.
        This article was recommended for Publication by Associate Editor T. L. Lam and Editor C. Yang upon evaluation of the revierwers' comments.
        This work was supported by the Deutsche Forschungsgemeinschaft (German Research Foundation) under grant number AL 1185/31-1.
        \textit{(Corresponding author: Matthias Mayer)}
        
        All authors are with the Technical University of Munich, TUM School of Computation, Information and Technology, Chair of Robotics, Artificial Intelligence and Real-time Systems, Boltzmannstrasse 3, 85748, Garching, Germany.
        {\tt\small $\{$\href{mailto:matthias.mayer@tum.de}{matthias.mayer}, \href{mailto:althoff@tum.de}{althoff}$\}$@tum.de}.

        Digital Object Identifier 10.1109/TASE.2025.3628162
    }
}

\newcommand{\authorShort}{
    \ifcensor
        {ANONYMOUS}%
    \else
        {MAYER, ALTHOFF}%
    \fi
}
\IEEEpubid{ \parbox{\textwidth-0.5in}{\vspace{40pt}\centering\copyrighttext}}

\begin{document}
\bstctlcite{IEEEexample:BSTcontrol}  %

\maketitle

\begin{abstract}
Modular robots have the potential to revolutionize automation, as one can optimize their composition for any given task. 
However, finding optimal compositions is non-trivial. 
In addition, different compositions require different base positions and trajectories to fully use the potential of modular robots. 
We address this problem holistically for the first time by jointly optimizing the composition, base placement, and trajectory to minimize the cycle time of a given task.
Our approach is evaluated on over $\SI{300}{}$ industrial benchmarks requiring point-to-point movements.
Overall, we reduce cycle time by up to $\SI{25}{\percent}$ and find feasible solutions in twice as many benchmarks compared to optimizing the module composition alone.
In the first real-world validation of modular robots optimized for point-to-point movement, we find that the optimized robot is successfully deployed in nine out of ten cases in less than an hour.

\textit{Note to Practitioners---}%
In industrial automation, there is a need for robots that adapt to specific tasks, thereby reducing cycle times and costs. 
Modular robots, which can be altered by rearranging building blocks similar to LEGO, offer a promising solution to this problem and are now available in industrial quality.
However, finding the optimal composition -- among the often more than a million conceivable options -- remains a challenge for humans, requiring automatic optimization.
This article presents a new method that starts from the 3D scan of the intended robot task and optimizes the final robot together with its position relative to the task and its program.
We focus on minimizing the cycle time of point-to-point movements, such as those required by a robot stocking a machine tool from a magazine or spot welding.
Empirically, our method works in almost all conducted experiments, achieving the promised cycle time.
The deployment is straightforward and only requires a few minutes of adapting the program within the graphical user interface provided by the robot manufacturer.
\end{abstract}

\begin{IEEEkeywords}
Modular robots,
robot programming,
motion and path planning, 
factory automation.
\end{IEEEkeywords}

\section{Introduction}
\label{sec:introduction}

\begin{figure}[t]
    \vspace*{0.3mm}  %
    \centering
    \import{figures/}{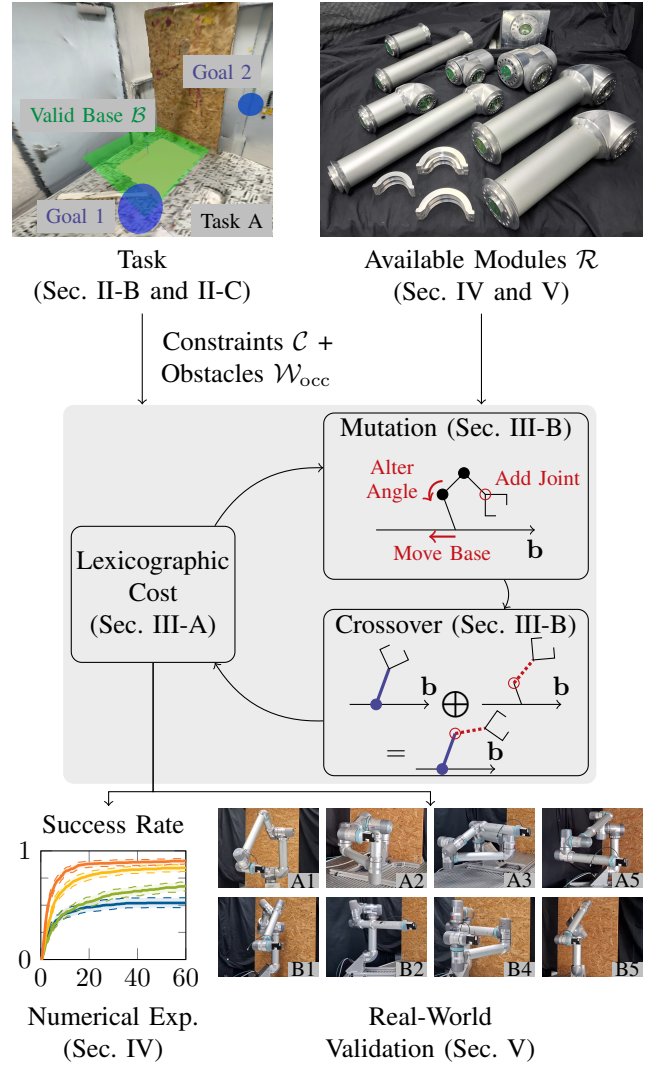}
    \caption{
    Overview of this article: The robot task and available modules $\modules$ are the inputs to our method, which optimizes modular robots.
    Our approach jointly optimizes the base of the robot, its module composition, and the trajectory to solve the task.
    Examples of results from simulations and real world experiments are shown last.
    }
    \label{fig:overview}
\end{figure}

\noindent \IEEEPARstart{M}{odular} reconfigurable robots have been a dream of roboticists to shape the future of automation\cite{Fukuda1988,Chocron1997a,Yang2000}.
One of their major promises is that task-specific kinematics can provide more efficient automation than standard industrial kinematics \cite[Ch.~22]{Siciliano2016},~\cite{Althoff2019,Liu2020}.
Their usage offers other benefits as well, such as a) economies of scale in producing standardized modules for these robots, b) easier deployment as they can be shipped and assembled from parts that can be handled manually, and c) easier maintenance as many different robots can be repaired with a limited set of distinct spare parts (that are again also easy to ship)\cite[Ch.~22.1]{Siciliano2016},~\cite{Yim2007,Liu2016}.
Those advantages are increasingly realized with hardware becoming readily available in industrial quality both from established companies such as Beckhoff\footnote{\label{fn:beckhoff}\href{https://www.beckhoff.com/de-de/produkte/motion/atro-automation-technology-for-robotics/}{beckhoff.com/de-de/produkte/motion/atro-automation-technology-for-robotics/}, \lastChecked}, as well as startups focusing on industrial\footnote{\label{fn:robco}\href{https://www.robco.de/en}{robco.de}, \lastChecked} or research\footnote{\label{fn:hebi}\href{https://www.hebirobotics.com/}{hebirobotics.com/}, \lastChecked} applications.

\IEEEpubidadjcol  %
Nonetheless, their application to point-to-point (PTP) movements -- the most common real-world automation task~\cite[p.~13]{IFR2024} -- as shown in \cref{fig:task_solutions}, has not yet been systematically evaluated in the real world (see the two rightmost columns of our related work summary in \cref{tab:overview_related_work}).
Implementation-wise, the variable kinematics of modular robots and the desire to find optimal ones for specific tasks, challenge the standard approaches for solving automation tasks~\cite[Ch.~22.4]{Siciliano2016}.
For example, the large number of conceivable robots -- a few thousand are often sensible (see \textit{design space} in \cref{tab:overview_related_work}) -- makes computational costs important.
Furthermore, changing kinematics challenge path planning, which is often tuned for specific robots \cite{MayerEffPathPlan}.

\paragraph{Contributions}
Our core contributions tackle these research gaps:
\begin{itemize}
    \item We implement a holistic optimization of a) modular robots, b) the placement of each robot relative to the task, and c) the trajectory required to solve point-to-point tasks.
    \item We systematically test this optimization in simulation\footnote{\label{fn:paper_webpage} Find all tasks, source code, and original data on the accompanying website \href{https://redirect.cps.cit.tum.de/hmro}{redirect.cps.cit.tum.de/hmro}. \lastChecked} and real-world experiments. 
\end{itemize}
Based on our experiments, we make the following claims:
\begin{itemize}
    \item Our holistic optimization converges quicker to better solutions \ref{hyp:better_convergence} and generalizes to varied tasks \ref{hyp:generalizes}.
    \item Optimizing cycle time also minimizes other costs \ref{hyp:other_costs}.
    \item Optimization results transfer to the real world with similar performance \ref{val:samePerformance} and minimal manual adaptations \ref{val:smallAdaptation}.
\end{itemize}

\paragraph{Organization}

The structure of this paper and its main method are shown in \cref{fig:overview}.
After summarizing related work, we formalize the optimization problem to be solved in \cref{sec:problemStatement}. 
Our solution is described in \cref{sec:implementation}. 
Lastly, we provide numerical optimization results in \cref{sec:num_example} and verify selected optimization results in real-world applications in \cref{sec:real_world_example}.

\subsection{Related Work}
\label{ssec:relatedWork}

\begin{table*}[t]  %
    \vspace{-1.7mm}
    \caption{Summary of Related Work}
    \label{tab:overview_related_work}
    \centering
    \begin{tabular}{cccccccc}
        \toprule
        Source & \multicolumn{3}{c}{Optimization Scope$^{\text{a}}$} & \multicolumn{2}{c}{Design Space} & Task Type$^{\text{b}}$ & Real \\
        \cmidrule(lr){2-4} \cmidrule(lr){5-6}
         & Mod. Robot & Base & Trajectory & Real $\reals^N$ & Disc. Choices & & Tests\\
        \midrule
        \cite{Karaman2011} & \xmark & \xmark & z & 0 & 0 & PP & 0 \\
        \cite{Kunz2013} & \xmark & \xmark & z & 0 & 0 & TG & 0 \\
        \cite{Pham2018} & \xmark & \xmark & z & 0 & 0 & TG & 0 \\
        \cite{MayerBPO} & \xmark & $\reals^3/\SEThree$ & (IK+z) & 3/6 & 0 & \textbf{PTP} & 0 \\
        \cite{MITSI200850} & \xmark & DH & IK & 4 & 0 & WSP & 0 \\
        \cite{Kim2021} & \xmark & $\reals \times [\SOTwo]^2$ & IK & 3 & 0 & MAN & 0 \\
        \cite{Baizid2010} & \xmark & $\reals^3$ & IK & 3 & 0 & \textbf{PTP} & 0 \\
        \cite{Alatartsev2015} & \xmark & \xmark & IK + z & 0 & 0 & \textbf{PTP} & 0 \\
        \midrule
        \cite{Althoff2019} & \checkmark & \xmark & (IK + z) & 0 & $\SI{32768}{}$ \cite{Klz2023OptimizingMR} & \textbf{PTP} & 0 \\
        \cite{Liu2020} & \checkmark & \xmark & (IK + z) & 0 & $\SI{167936}{}$ & WST/\textbf{PTP}/MAN & 0 \\
        \cite{Klz2023OptimizingMR} & \checkmark & \xmark & (IK + z) & 0 & $\approx 10^{12}$ & \textbf{PTP} & 0 \\
        \cite{Whitman2020} & \checkmark & \xmark & (IK) & 0 & $\approx \SI{1e6}{}$ \cite{Klz2023OptimizingMR} & WSP & 1 \\
        \cite{Icer2017} & \checkmark & \xmark & (IK + z) & 0 & $\SI{15552}{}$ \cite{Klz2023OptimizingMR} & WSP & 0 \\
        \cite{Ha2018} & \checkmark{} + CL & \xmark & (IK) & 6 & $\gtrapprox \SI{15625}{}$ & WST & 2 \\
        \cite{Romiti2023} & \checkmark & $\SETwo$ & (IK) & 0 & $\approx \SI{2.2e+05}{}$ & WSP & 4 \\
        \cite{Lei2024} & \checkmark & $\SETwo$ & (IK) & $\SI{3}{}$ & $\approx \SI{2.9e+05}{}$ & WST & 3 \\
        \cite{Baumgartner2024} & \xmark{} [CL] & $\reals^4$ & IK & $\SI{22}{}$ & 123 & WSP & 0 \\
        \cite{Hoffman2025} & \checkmark (+ CL) & ($\reals \times \SOTwo$) & (IK) & $(\SI{1058}{})$ & $\SI{2187}{}$ & WSP & 0 \\
        \cite{Kulz2024} & \checkmark & (Mobile) & (IK+z) & 0 & $\SI{30005}{}$ & WST & 6 \\
        \midrule
        Our & \checkmark & $\reals^3/\SETwo$ & IK+(z) & $\SI{3}{}$ & $\gtrapprox \SI{2.3e+15}{}$ & \textbf{PTP} & \textbf{10} \\
        \bottomrule\\
        \multicolumn{8}{p{5.5in}}{
            \textbf{\checkmark}:~Considered, \textbf{\xmark}:~Not considered, \textbf{+}:~Combination, \textbf{/}:~Both considered
        }\\
        \multicolumn{8}{p{5.5in}}{
            $^{\text{a}}$\underline{\smash{Scope}} \, \textbf{(in bracket)}:~Solved separately for each robot design, \textbf{M}:~Modular robot, \textbf{CL}:~Continous link parameters, \textbf{DH}:~Denavit Hartenberg parameters, \textbf{IK}:~Find individual inverse kinematics, \textbf{z}:~Find trajectory
        }\\
        \multicolumn{8}{p{5.5in}}{
            $^{\text{b}}$\underline{\smash{Task Type}}\, \textbf{MAN} - Manipulability in part of the workspace, \textbf{PP}:~Path planning, \textbf{PTP}:~Point-to-point movement, \textbf{TG}:~Trajectory generation, \textbf{WSP}:~IK solved at work space poses, \textbf{WST}:~Work-space trajectory
        }\\
    \end{tabular}
\end{table*}

\noindent Our literature overview surveys methods for optimizing a) modular robots, b) base poses, and c) planned motions; we also survey common robotic tasks in the manufacturing industry.
We summarize our survey in \cref{tab:overview_related_work}, highlighting which joint optimizations have been considered in the literature and how they compare to this work.

\subsubsection{Modular Robot Optimization}
\label{ssec:relWork:MRO}

A key motivation for this article is the efficient utilization of reconfigurable robots~\cite[Ch. 22.2]{Siciliano2016}, where even small module sets can be used to assemble millions of possible robots (see \cref{tab:overview_related_work}, column \textit{design space}). 
Recent solutions to find optimal assemblies have combined hierarchical elimination with kinematic restrictions~\cite{Liu2020,Althoff2019}, genetic algorithms~\cite{Icer2017,Klz2023OptimizingMR,Romiti2023,Lei2024,Kulz2024,Hoffman2025,Yang2000}, heuristic search~\cite{Ha2018}, or reinforcement learning~\cite{Whitman2020}.
Similar problems have to be solved within soft robotics \cite{Stroppa2024,Stroppa2025}, and continuous link (CL) optimization \cite{Baumgartner2024,Ha2018,Hoffman2025}.

Previous works considered various task types (see \cref{tab:overview_related_work}), e.g., entailing a list of specific workspace poses (WSP)~\cite[Case 2]{Liu2020},~\cite{Whitman2020, Althoff2019,Yang2000}, pre-defined work space trajectories~(WST) to follow~\cite[Case 1]{Liu2020},~\cite{Ha2018,Romiti2023,Lei2024,Baumgartner2024,Kulz2024}, or maximizing robot manipulability in a certain area (MAN)~\cite[Case 3]{Liu2020}, \cite{Hoffman2025}.
Required solution fidelity ranges from the existence of inverse kinematic solutions for desired end-effector poses (WSP)~\cite{Whitman2020} to physically feasible trajectories (PTP, WST), e.g.,~\cite{Liu2020}.

Their advantages, e.g., regarding cycle time, compared to standard manipulators are shown in \cite{Liu2020,Althoff2019}.
Some papers also test the optimization results on real hardware~\cite{Ha2018,Romiti2023,Lei2024,Whitman2020,Kulz2024}.

\subsubsection{Base Pose Optimization}

Previously published work on optimizing the base pose of a robot often uses black-box optimization algorithms, such as genetic algorithms~\cite{MITSI200850} and Bayesian optimization~\cite{Kim2021}. 
Other optimizers, such as grid search~\cite{Lechler2021}, or gradient-based methods~\cite{Son2019} have also been applied.
These approaches were compared in~\cite{MayerBPO}.
Many of these works also highlight that robot behavior, such as the used inverse kinematic solutions, should be optimized jointly with the base of the robot~\cite{MITSI200850,Kim2021,Baizid2010}, which for changing robots has only been done in \cite{Baumgartner2024} (indicated by parentheses in \textit{Trajectory} column of \cref{tab:overview_related_work}).

The first steps towards joint optimization of the base pose of a (modular) robot and its configuration have been taken in~\cite{Romiti2023,Lei2024,Baumgartner2024,Hoffman2025,Kulz2024}.
The works in \cite{Romiti2023,Lei2024} use genetic algorithms to optimize a robot module configuration and its mounting position for a single trajectory tracking task and demonstrate this in real-world experiments.
In contrast, \cite{Baumgartner2024} considers continuous robot parameters, which enables the joint optimization of the trajectory and the length of the robot links (treating them as joints with zero velocity) via collocation.
As collocation is fast, they can find the best  combination of prismatic and revolute joints by enumerating all combinations resulting in robots with six degrees of freedom.
The work in \cite{Hoffman2025} optimizes discrete joint choices in an outer loop with ant colony optimization and, in an inner loop, optimizes the base, link lengths, and inverse kinematic~(IK) solutions.

\subsubsection{Optimal Path and Task Planning}

In general, robotic path and task planning (PP, PTP in \cref{tab:overview_related_work}) is a hard problem to solve optimally due to a) its complex state space~\cite[Ch. 7]{Siciliano2016} and b) the problem that IK typically has non-unique solutions to desired workspace goals~\cite{Alatartsev2015}.
The work in~\cite{Baizid2010} solved this by encoding IK solutions of goals in genes optimized by a genetic algorithm.

Common asymptotically optimal path planning methods are sampling-based planners that incrementally improve solutions, such as RRT*/PRM*~\cite{Karaman2011}.
One derivative is Lazy-PRM*\footnote{\label{fn:LazyPRMStar}\url{https://ompl.kavrakilab.org/classompl_1_1geometric_1_1LazyPRMstar.html}, \lastChecked} implemented by OMPL~\cite{Sucan2012}, which can a) keep planning information during multiple evaluations of the same robot, b) is asymptotically optimal~\cite{Karaman2011}, and c) utilizes lazy collision checking to accelerate road map construction~\cite{Bohlin2000}.

Specific adaptations for more efficient path planning under changing kinematic structures of modular robots have been published in~\cite{MayerEffPathPlan,Baumgartner2024}.
Either one reuses paths found on previously considered robots and repairs them to fit the new kinematics and collision geometries of similar ones \cite{MayerEffPathPlan}, or one optimizes the path in joint space jointly with the robot parameters \cite{Baumgartner2024}.

Some works, such as \cite{Althoff2019,Liu2020,Klz2023OptimizingMR,Kulz2024} labeled with the PTP \textit{task type} in \cref{tab:overview_related_work}, additionally use trajectory generation (TG) to judge the performance of the optimized robots with regard to, e.g., cycle time.
In \cite{Althoff2019}, trapezoidal velocity profiles (TVP) were used, and \cite{Liu2020,Klz2023OptimizingMR,Kulz2024} used \cite{Kunz2013}, which adds circular blends at via-points.
Both TG methods (TVP, \cite{Kunz2013}) only support fixed velocity and acceleration limits, which might not completely utilize the available joint torques.
Newer TG methods, such as \cite{Pham2018} employing reachability analysis, have not yet been applied in modular robot optimization, though they can enforce user-defined torque limits.
This additional optimization problem is avoided if an end-effector trajectory is given as the task (WST), such as done by~\cite{Lei2024,Ha2018}, or if only point-wise inverse kinematic solutions (WSP) are required~\cite{Icer2017,Romiti2023,Whitman2020,Baumgartner2024,Hoffman2025}.

\subsubsection{Task Description}

A recently updated overview of domain-specific languages for robotics\footnote{\url{https://corlab.github.io/dslzoo/all.html}, \lastChecked} was first published in~\cite{Nordmann2016}, but it does not include a unified framework to describe (modular) robots and industrial tasks.
Such a unified framework has been published as the Composable Benchmark for Robotics Applications (CoBRA)~\cite{CoBRA}.
It contains baselines for module optimization with genetic algorithms~\cite{Klz2023OptimizingMR} and various base-optimization algorithms~\cite{MayerBPO}, solving industrial point-to-point (PTP) tasks in static environments.
These PTP tasks are, according to~\cite[p.~13]{IFR2024}, still the majority ($\SI{77.6}{\percent}$) of industrial robotic tasks.
Furthermore, a low number of deployed cobots ($\qty{10}{\percent}$ market share, \cite[p.~15]{IFR2024}) indicates that most robotic tasks are still within static environments.
Modular robots have been optimized for different cost functions, e.g., energy consumption~\cite{Liu2020}, robot mass as a proxy of complexity~\cite{Whitman2020}, the trajectory tracking error~\cite{Baumgartner2024}, or cycle time~\cite{Klz2023OptimizingMR}.

In summary and as highlighted in \cref{tab:overview_related_work}, there is a lack of research on the major class of industrial robotic tasks -- point-to-point (PTP) movements -- and how to optimize modular robots for these.
Furthermore, systematic real-world validations of modular robots optimized for PTP have not yet been done.
Combining these three is our main contribution.

\section{Task Description and Problem Statement}
\label{sec:problemStatement}

\noindent This section defines the optimization problem considered in this paper, following the notation used in \cite{CoBRA}.
We denote scalars with lowercase letters, vectors with bold lowercase letters, matrices with uppercase and bold letters, and sets with calligraphic letters.

The $i^\mathrm{th}$ element of a vector $\Jposv$ is denoted by $\Jposv_i$, and comparisons of vectors are performed elementwise $\Jposv_a \leq \Jposv_b \Leftrightarrow \bigwedge_{i=1}^n q_{a, i} \leq q_{b, i}$.
$\Zerov_n = [0, \ldots, 0]^T \in \reals^n$ is the vector of $n$ zeros.
The Euclidean vector norm is denoted by $\norm{\Transv} = \sqrt{\sum_{t \in \Transv} t^2} \in \reals$, $\normElem{\Transv} = \left[ \normElem{t_1}, \ldots, \normElem{t_n} \right]^T \in \reals^n$ is the elementwise absolute value, and $\normElem{\G} \in \naturals$ is the cardinality of a set which returns the number of elements in the set $\G$.
We denote integer intervals as $\naturalInt{a} = \left\{ x \in \naturals \mid 1 \leq x \leq a \right\}$, and real intervals by $\realInt{a}{b} = \left\{ x \in \reals \mid a \leq x \leq b \right\}$.
Lastly, we define the indicator function mapping Boolean values/expressions $\bools$ to $\{0, 1\}$:

\begin{equation}
    \label{eq:indicator}
    \indicatorF{b} = \begin{cases}
        1, &\mathrm{if} \, b, \\
        0, &\otherwise.
    \end{cases}
\end{equation}

Our default representation for any \textit{pose} is the homogeneous transformation $\T \in \SEThree$ as defined in \cite[Sec.~3.3.1]{Lynch2017}.
In summary, it combines the rotation matrix $\rotM \in \SOThree \subset \reals^{3 \times 3}$ and translation vector $\Transv \in \reals^3$ between two coordinate systems $A$ and $B$:

\begin{equation}
    \T^A_B = \begin{bmatrix}
        \rotM^A_B & \Transv^A_B\\
        \Zerov_{3}^T & 1
    \end{bmatrix}.
\end{equation}
The matrix $\T^A_B$ can be applied to any vector $\Pointv_A \in \reals^3$ in the coordinate system $A$ to represent it in the coordinate system $B$, by concatenating $\Pointv_A$ with a one:

\begin{equation}
    \begin{bmatrix}
        \Pointv_B\\
        1
    \end{bmatrix} = \T^A_B \begin{bmatrix}
        \Pointv_A\\
        1
    \end{bmatrix} = \begin{bmatrix}
        \rotM^A_B \Pointv_A + \Transv^A_B \\
        1
    \end{bmatrix}.
\end{equation}
Similar matrices exist in 2D, i.e., $\SOTwo \subset \reals^{2 \times 2}$ for rotation in a plane and $\SETwo$ for planar translation and rotation.

Within this paper, we omit the homogeneous coordinate if unambiguous.
We use $\Transv(\T)$ to denote the translation vector inside $\T$ and the shorthand $\norm{\T} = \norm{\Transv(\T)}$ to denote the Euclidean distance of the pose from the origin.
Similarly, $\rotM(\T)$ denotes the rotation matrix inside $\T$, and the function $\theta(\T)$ returns the angle from the angle-axis representation of $\rotM(\T)$ \cite[Tab.~2.2]{Siciliano2016}.

\subsection{Robot Model}
\label{ssec:problem:robotmodel}

\noindent We assume the model of a stiff, modular robotic manipulator with $\dof$ joints connected by $\dof+1$ links $\link_i$.
Its state is given by joint positions $\Jposv \in \reals^{\dof}$ and velocities $\Jvelv \in \reals^{\dof}$.
Torques $\Jtorqv \in \reals^{\dof}$ and external wrenches $\Fext$ result in joint accelerations $\Jaccv \in \reals^{\dof}$.
The robot is built from $\nModules$ modules $\module{i}$ from a set of module types $\modules$; in general, the same type can be used multiple times.
Each robot is uniquely described by the assembled modules $\assembly = [\module{1}, \ldots, \module{\nModules}],\, \module{i} \in \modules$ listed from the base to the end effector, and the pose of the robot base $\B \in \SEThree$.
With these, we can define the

\begin{itemize}
    \item forward kinematics that returns the end effector pose $\Teff(\Jposv, \assembly) \in \SEThree$ relative to the base $\B$;
    \item occupancy of any link $\occupancy_{\link_i}(\Jposv, \B, \assembly) \subset \powerset(\reals^3)$, where $\powerset(\bullet)$ returns the power set of $\bullet$;
    \item robot occupancy $\occupancy(\Jposv, \B, \assembly) = \bigcup\limits_{i=1}^{\dof + 1} \occupancy_{\link_i}(\Jposv, \B, \assembly)$; 
    \item forward dynamics: $\Jaccv = \dynamics(\Jposv, \Jvelv, \Jtorqv, \Fext, \B, \assembly)$;
    \item inverse dynamics: $\Jtorqv = \dynamics^{-1}(\Jposv, \Jvelv, \Jaccv, \Fext, \B, \assembly)$.
\end{itemize}

\subsection{Hybrid Motion Planning Problem}
\label{ssec:hybMotPlan}

\noindent Following \cite{CoBRA}, we consider any robotic task defined by a set of constraint functions $\constraints =\{\constraint_1, \ldots, \constraint_{\normElem{\constraints}}\}$ (see \cref{ssec:Constraint}).
A solution to such a task requires a
\begin{itemize}
\item robot assembly $\assembly$,
\item base pose $\B \in \SEThree$, and
\item desired state-input vector $\Jmotionv(t) = \left[ \Jposv_\desired(\timenow), \Jvelv_\desired(\timenow), \Jaccv_\desired(\timenow) \right]^T$ containing the desired robot joint positions $\Jposv_\desired(\timenow)$, velocities $\Jvelv_\desired(\timenow)$, and accelerations $\Jaccv_\desired(\timenow)$ during the time interval $\realInt{0}{\timeend}$ required to solve the task (without loss of generality, we set $t_0 = \timezero$).
\end{itemize}

To find optimal solutions, we minimize any cost function $\costfun{\costid}$ consisting of terminal costs $\Phi_\costid$ and the integral of running costs $L_\costid$:

\begin{multline}
    \label{eq:generalCostFunction}
    \costfun{\costid}(\Jmotionv, \B, \assembly) = 
    \Phi_\costid(\Jmotionv(\timezero), \Jmotionv(\timeend), \timeend, \B, \assembly) + \\
    \int_{\timezero}^{\timeend} L_\costid(\Jmotionv(\timeany), \timeany, \B, \assembly) \mathrm{d}\timeany.
\end{multline}

Formally, we define the \textit{hybrid motion planning problem} as finding a module assembly $\assembly^*$, base placement $\B^*$, and desired state-input vector $\Jmotionv^*$ minimizing the cost function $\costfun{\costid}$:
\begin{equation}
    [\assembly^*, \B^*, \Jmotionv^*] = 
    \argmin_{\assembly, \B, \Jmotionv} \costfun{\costid}(\Jmotionv(t), \timeend, \B, \assembly)
    \label{equ:hybOptim}
\end{equation}
subject to $\forall \timenow \in \realInt{\timezero}{\timeend}$:
\begin{align} %
    \Jaccv_\desired &= \dynamics(\Jposv_\desired, \Jvelv_\desired, \Jtorqv_\desired, \Fext, \B, \assembly) \\
    \forall \constraint \in \constraints \colon \constraint (\Jmotionv, \timenow, \B, \assembly) &\leq 0.
\end{align}
Subsequently, we introduce our definitions for constraints in $\constraints$.

\subsection{Constraints}
\label{ssec:Constraint}

\noindent Constraints define the goals the robot should achieve and how they should be solved to be physically feasible.
A constraint function 
\begin{equation}
    \constraint \colon \Jmotionv \times \timenow \times \B \times \assembly \to \reals
\end{equation}
can capture all limitations we require and is satisfied if it is negative or zero.
Within this work, we enforce the following robotic constraints:

\begin{itemize}
    \item No self-collisions between links $\link_i$ of the robot:
    \begin{equation}
        \forall i, j \in \naturalInt{\dof + 1}, i \neq j \colon \occupancy_{\link_i} \cap \occupancy_{\link_j} = \emptyset
        \label{eq:selfCollFree}
    \end{equation}
    \item No collisions with obstacles occupying $\obstacle \subset \powerset(\reals^3)$:
    \begin{equation}
        \occupancy(\Jposv_\desired, \B, \assembly) \cap \obstacle = \emptyset
        \label{eq:collFree}
    \end{equation}
    \item State and input constraints: 
    \begin{equation}
        \Jposvmin \leq \Jposv_\desired \leq \Jposvmax \wedge \normElem{\Jvelv_\desired} \leq \Jvelvmax \wedge \normElem{\Jaccv_\desired} \leq \Jaccvmax
        \label{eq:kinLimit}
    \end{equation}
    \item Torque limits:
    \begin{equation}
        \normElem{\Jtorqv_\desired} = \normElem{\dynamics^{-1}(\Jmotionv, \Fext, \B, \assembly)} \leq \Jtorqvmax 
        \label{eq:torqueLimit}
    \end{equation}
    \item All goals $\G$ are fulfilled:
    \begin{equation}
        \label{eq:constraint:goalFulfilled}
        \forall \goal[] \in \G \, \exists t_{\goal[]} \in \realInt{\timezero}{\timeend} \colon \goal(\Tdes[\goal], \Jmotionv(t_{\goal[]}))\leq 0
    \end{equation}
    \item Goals are fulfilled in order:
    \begin{equation}
        \label{eq:constraint:goalOrder}
        \forall \goal[i], \goal[j] \in \G,  i < j \colon t_{\goal[i]} < t_{\goal[j]}
    \end{equation}
    \item Base pose limits: $\B \in \validBases$.
\end{itemize}

\textit{Goals} $\goal$ are a special type of constraint that need to be fulfilled at one time $t$ by the robot \cref{eq:constraint:goalFulfilled}.
We note that all task types in \cref{tab:overview_related_work} have been formalized in \cite[Tab. IV]{CoBRA}.

Focusing on point-to-point movements, we define a single goal requiring the robot to \textit{reach} a set of poses in the workspace \cite{CoBRA}.
The desired set is given by a desired (nominal) pose $\Tdes[\goal] \in \SEThree$ and a certain tolerance.
We define the tolerances by $i \in \naturalInt{n}$ projections 
\begin{equation}
    \label{eq:projection}
    s_i \colon \SEThree \to \reals.
\end{equation} 
The tolerance is fulfilled if all projected values $s_i(\T)$ are inside their respective interval $\realInt{\Tolbound_{i, \min}}{\Tolbound_{i, \max}}$.
A typical set of projections are the Cartesian coordinates $\Proj = [x, y, z]$ which can constrain the end-effector position to a cube of width $w > 0$ around the desired pose $\Tdes[\goal]$ with the interval $\realInt{-\sfrac{w}{2}}{\sfrac{w}{2}}$ for each direction.
Additionally, stopping is ensured by reaching velocities $\Jvelv_\desired$ and accelerations $\Jaccv_\desired$ below a small $\epsilon$.
Formally, a reach goal $\goal$ is therefore fulfilled at time $\timenow$ if

\begin{multline}
    \goal(\Tdes[\goal], \Jmotionv(\timenow)) = \|\Jvelv_\desired(\timenow)\|_2 \leq \epsilon \wedge \|\Jaccv_\desired(\timenow)\|_2 \leq \epsilon \,\wedge \\ 
    \forall i \in \naturalInt{n} \colon s_i(\Tdes[\goal]^{-1} \, \B \, \Teff(\Jposv_\desired(\timenow),  \assembly)) \in \realInt{\Tolbound_{i, \min}}{\Tolbound_{i, \max}}.
\end{multline}

We note that the solution to each reach goal $g$ is usually non-unique, i.e., there exists an (infinite) set of inverse kinematic solutions $\IKsols_g = \left\{ \Jposv_{g} \mid \goal \left( \Tdes[\goal], \left[ \Jposv_{g}, \Zerov, \Zerov \right]^T \right) \right\}$.

\section{Method}
\label{sec:implementation}

\noindent We jointly optimize the composition, base pose, and trajectory using hierarchical elimination with a lexicographic cost function, as introduced next.
Afterwards, we show how a solution of this joint optimization can be encoded by a single genome and how the genetic operators for the genetic algorithm are defined.

\subsection{Hierarchical Elimination}
\label{ssec:hierElim}

\noindent To evaluate the main objective \cref{eq:generalCostFunction}, one needs to know how the robot moves, i.e., find $\Jmotionv$.
We can construct $\Jmotionv$ or reject its existence with steps of increasing computational complexity, which was termed \textit{hierarchical elimination} by \cite{Icer2016}.
For example, it is much faster to add up the length of the robot links and compare this maximum reachable distance with the distance from the robot base to all goals than it is to find IK solutions for all goals.
If the robot is too short for at least one goal, there is no need to find IK solutions for it.

Feedback from these hierarchical steps can be integrated into genetic algorithms either by a) removing all individuals failing a single step from consideration~\cite{Icer2017}, b) penalizing any failed step with a ``failure cost''~\cite{MayerBPO}, or c) lexicographic cost functions such as applied by~\cite{Klz2023OptimizingMR}. 
From these three, lexicographic costs outperformed simpler cost functions in~\cite{Klz2023OptimizingMR} as they give the most fine-grained feedback to the optimization.

Therefore, we use the lexicographic cost function from~\cite[Sec.~III.B]{Klz2023OptimizingMR} and generalize it by adding the base $\B$ and trajectory $\Jmotionv$ as arguments to the cost function, which fits our extended optimization scope.
Overall, we define a lexicographic cost as a sequence of $n$ costs ordered by increasing importance and computational complexity 
\begin{equation}
    \costfun{}(\Jmotionv, \B, \assembly) = \left[ \costfun{1}(\Jmotionv, \B, \assembly), \ldots, \costfun{n}(\Jmotionv, \B, \assembly) \right].
\end{equation}
It will become evident in \cref{eq:sol:J} that this returns the same final costs as \cref{eq:generalCostFunction}.
These lexicographic costs can then be ordered, as needed, e.g., by the selection process in genetic algorithms:
\begin{align}
\begin{split}
    \costfun{}(\Jmotionv[a], \B_a, \assembly_a) > \costfun{}(\Jmotionv[b], \B_b, \assembly_b) &\iff \\
    \exists k \in \naturalInt{n} \colon \costfun{k}(\Jmotionv[a], \B_a, \assembly_a) &> \costfun{k}(\Jmotionv[b], \B_b, \assembly_b) \, \wedge \\
    \forall i < k \colon \costfun{i}(\Jmotionv[a], \B_a, \assembly_a) &= \costfun{i}(\Jmotionv[b], \B_b, \assembly_b),
\end{split}\\
\begin{split}
     \costfun{}(\Jmotionv[a], \B_a, \assembly_a) = \costfun{}(\Jmotionv[b], \B_b, \assembly_b) &\iff \\
     \forall i \in [1, n] \colon \costfun{i}(\Jmotionv[a], \B_a, \assembly_a) &= \costfun{i}(\Jmotionv[b], \B_b, \assembly_b).
\end{split} 
\end{align}

As an example, consider three robots $R_1, R_2, R_3$ and two cost functions: $J_1$ is $0$ if the length of a robot $R$ is longer than the distance to the furthest goal and $1$ otherwise, and $J_2$ counts how many goals have no valid IK solution.
$R_1$ is too short, i.e., $J_1(R_1) = 1$.
The other two robots $R_2, R_3$ are long enough so $J_1(R_2) = J_1(R_3) = 0$.
Nevertheless, no IK solution is found for one goal with $R_2$, so $J_2(R_2) = 1$ and $J_2(R_3) = 0$.
Ordering $J_1$ and $J_2$ lexicographically, i.e., $J = [ J_1, J_2 ]$, we obtain $J(R_1) = [1, x] > J(R_2) = [0, 1] > J(R_3) = [0, 0]$.
This order is, e.g., used by the selection operator of the genetic algorithm to prefer the better $R_3$ as a parent for the next generation.
Note that the evaluation of $J_2(R_1)$ can be skipped, as $\forall x \in \reals \colon J(R_1) > J(R_2) > J(R_3)$.

Next, we present our considered cost terms, extending the previous state of the art \cite[Eq.10-13]{Klz2023OptimizingMR}.
We shorten the notation by the novel use of \textit{recursive} lexicographic costs; i.e., the cost terms \cref{eq:simpleIK:J,eq:path:J,eq:traj:J,eq:sol:J} are themselves lexicographic.
Leveraging hierarchical elimination, if the calculation of a cost fails, the following ones are skipped:

\begin{lexicographic}
    \item The \textbf{robot length} cost judges whether the maximum Euclidean distance between any goal and the base $\max\limits_{\goal \in \G} \norm{\Tdes[\goal]^{-1} \B}$ is smaller than the maximum length of the robot and terminates if the robot is too short.
    The robot length is over-approximated by the sum of module lengths $d(\module{i}$) (generalization of~\cite[Eq.~10]{Klz2023OptimizingMR}):
    \begin{equation}
        \label{eq:robLengthCost}
        \costfun{\thelexicographici}(\assembly, \B) = \indicatorF{\sum\limits_{\module{i} \in \assembly} d(\module{i}) > \max\limits_{\goal \in \G} \norm{\Tdes[\goal]^{-1} \B}}.
    \end{equation}
    \item The \textbf{module available} cost checks whether the required modules are available and terminates the evaluation if any module is missing.
    Using $n_{\mathrm{avail}} \colon \modules \to \naturals$, which returns the number of modules of type $\module{} \in \modules$ available, the cost counts the number of missing modules:
    \begin{equation}
        \label{eq:cost:availableModules}
        \costfun{\thelexicographici}(\assembly) = \sum_{m' \in \modules} \max \left( 0, \sum_{\module{} \in \assembly} \indicatorF{\module{} = m'} - n_{\mathrm{avail}}(m') \right).
    \end{equation}
    \item \label{cost:ik} The \textbf{robot inverse kinematic (IK) without obstacles} cost tests how many goals $\goal \in \G$ have a valid inverse kinematic solution and returns the sum of residual distances to the desired end-effector poses $\Tdes[\goal]$.
    For each goal $\goal \in \G$, we try to determine IK solutions $\Jposv_{\goal}$ numerically with a maximum of $\nSimpleIK$ steps.
    Even if no solution fulfilling $\goal$ is found, we still return the best found $\Jposv_{\goal}$ that minimizes a weighted sum $d$ of translational and rotational distance, i.e.,
    \begin{equation}
    \label{eq:cost:ik:distance}
    \begin{split}
        d(\Tdes[\goal], \Teff(\Jposv_{\goal}, \assembly)) &= \norm{\Tdes[\goal]^{-1} \Teff(\Jposv_{\goal}, \assembly)} + \\
            & \quad w \,\theta(\Tdes[\goal]^{-1} \Teff(\Jposv_{\goal}, \assembly)).
    \end{split}
    \end{equation}
    By combining \cref{eq:cost:ik:distance} and \cite[Eq.~11]{Klz2023OptimizingMR}, we obtain
    \begin{equation}
    \label{eq:simpleIK:J}
    \begin{split}
        \costfun{\thelexicographici}(\assembly, \B) = \Biggl[ \Biggr. & -\sum_{\goal \in \G} \indicatorF{\exists \Jposv_{\goal} \colon \goal \left( \Tdes[\goal],
        \left[
            \Jposv_{\goal}, \Zerov, \Zerov
        \right]^T
        \right)
        }, \\
        &\sum_{\goal \in \G} d \left( \Tdes[\goal], \Teff(\Jposv_{\goal}, \assembly) \right) \Biggl. \Biggr] .
    \end{split}
    \end{equation}
    \item The \textbf{Robot IK with obstacles} cost $\costfun{\thelexicographici}(\assembly, \B)$ is the same as $\cref{eq:simpleIK:J}$.
    The only difference is that the IK search continues if $\Jposv_{\goal}$ violates \cref{eq:collFree,eq:selfCollFree}, i.e., the robot is in a (self-) collision.
    This problem is harder to solve and therefore may use additional iterations $\nCplxIK$.
    \item The \textbf{Robot IK joint limits} cost checks whether the found IK solutions respect the joint limits of the robot, especially the maximum joint torques $\Jtorqvmax$.
    The cost counts the number of invalid IK solutions:
    \begin{equation}
        \costfun{\thelexicographici}(\assembly, \B) = \sum_{\goal[] \in \G} \indicatorF{\Jposv_{\goal} \, \mathrm{violates} \,\cref{eq:kinLimit} \vee \cref{eq:torqueLimit}} 
    \end{equation}
    \item \label{cost:path_planning} Following previous work~\cite{Liu2020,Klz2023OptimizingMR}, we use \textbf{path planning} to determine whether consecutive goals $\goal[i], \goal[i+1]$ with IK solutions $\Jposv_{\goal[i]}, \Jposv_{\goal[i+1]}$ are connectable with collision-free paths $\qpath_i(s)$ and how long these paths are.
    We assume that each path is piece-wise linear and consists of $n_i$ line segments $\linesegv_{i, j} \colon \realInt{0}{1} \to \reals^{\dof}$ that connect, i.e., $\forall j \in \naturalInt{n_i - 1} \colon \linesegv_{i, j}(1) = \linesegv_{i, j+1}(0)$.
    Formally, the path is $\qpath \colon \realInt{0}{n_i} \to \reals^{\dof}, \qpath_i(s) = \linesegv_{i, \lfloor s \rfloor}(s - \lfloor s \rfloor)$.
    
    Each path $\qpath_i(s)$ is planned with an anytime planner with a time limit of $\tplan$ seconds.
    If the planner succeeds, it finds a path free of (self-)collisions:
    \begin{equation}
    \label{eq:path:collfree}
    \begin{split}
        &\forall s \in \realInt{0}{n_i} \colon \qpath_i(s) \,\mathrm{satisfies} \,\cref{eq:selfCollFree} \wedge \cref{eq:collFree},
    \end{split}
    \end{equation}
    and the ends of the path are determined by the previously found IK solutions, i.e., 
    \begin{equation}
    \label{eq:path:ends}
        \left( \qpath_i(0) = \Jposv_{\goal[i]} \right) \wedge \left( \qpath_i(n_i) = \Jposv_{\goal[i+1]} \right).
    \end{equation}
    
    An underapproximation of the path length is determined by the time it would take to track each line segment of $\qpath_i$ with constant maximum joint velocity $\Jvelvmax$:
    \begin{equation}
        f_{\mathrm{t}}(\qpath_i(s)) = \sum_{s'=0}^{n_i} \max_{j \in \naturalInt{\dof}} \frac{\normElem{\qpath_i(s'+1) - \qpath_i(s')}_j}{\Jvelvmax[j]}
    \end{equation}
    The final cost is
    \begin{equation}
    \label{eq:path:J}
    \begin{split}
        \costfun{\thelexicographici}(\assembly, \B) = \Biggl[ \Biggr. & -\sum_{i = 1}^{\normElem{\G} - 1} \indicatorF{ \exists \qpath_i(s) \colon \cref{eq:path:collfree} \wedge \cref{eq:path:ends} }, \\
            & \sum\limits_{i = 1}^{\normElem{\G} - 1} f_{\mathrm{t}}\left(
                \qpath_i(s)
            \right)
        \Biggl. \Biggr].
    \end{split}
    \end{equation}
    \item Within \label{cost:traj_param} \textbf{trajectory generation}, we determine a time parameterization for each previously found path $\qpath_i$, which can fail, e.g., due to unsatisfiable torque requirements.
    If successful, trajectory generation finds a smooth function $\Jposv_{i} \colon \realInt{\timezero}{\timeend[i]} \to \reals^{\dof}$ tracking each path $\qpath_{i}(s)$ within time $\timeend[i]$.
    The generated trajectory $\Jposv_i(t)$ explicitly respects the joint limits in a) position $\Jposvmin, \Jposvmax$, b) velocity $\Jvelvmax$ \cref{eq:kinLimit}, and c) torque $\Jtorqvmax$ \cref{eq:torqueLimit}.
    Furthermore, it stays within a given maximum deviation $\delta \geq 0$ of the path $\qpath_i(s)$:
    \begin{equation}
    \begin{split}
        \label{eq:traj:delta}
        \forall \timenow \in \realInt{\timezero}{\timeend[i]} \exists s \in \realInt{0}{n_i} \colon \norm{\Jposv_i(\timenow) - \qpath_i(s)} \leq \delta.
    \end{split}
    \end{equation}
    
    Here, the cost is the number of successful parameterizations and their combined execution time:
    \begin{equation}
    \label{eq:traj:J}
    \begin{split}
        \costfun{\thelexicographici} = \Biggl[ \Biggl. & -\sum_{i = 1}^{\normElem{\G} - 1} \indicatorF{ \exists \Jposv_{i}(t) \colon \cref{eq:kinLimit} \, \wedge \, \cref{eq:torqueLimit} \wedge \cref{eq:traj:delta} }, \\
        & \sum \limits_{i = 1}^{\normElem{\G} - 1} \timeend[i] \Biggl. \Biggr]
    \end{split}
    \end{equation}
    \item \label{cost:solution} Finally, we create the candidate \textbf{solution} by concatenating the previously created trajectories $\Jposv_{i}(t)$ to form $\Jmotionv(t)$.
    With this, we can calculate the final costs $\costfun{\costid}$ for solving a task using \cref{eq:generalCostFunction}.
    In addition, the number of failed constraints and goals is returned, which can happen, e.g., due to the allowed deviation of the trajectory $\delta$ or constraints not explicitly handled by previous steps:
    \begin{equation}
    \label{eq:sol:J}
    \begin{split}
        \costfun{\thelexicographici} = \Biggl[ \Biggl. &\sum_{\constraint \in \constraints} \indicatorF{\exists \timenow \colon \constraint(\Jmotionv(t), \timenow, \B, \assembly) > 0 }, \\
        & \sum_{\goal \in \G} \indicatorF{ \nexists \timenow \colon \goal(\Tdes[\goal], \Jmotionv(t)) }, \\
        & \costfun{\costid}(\Jmotionv, \B, \assembly) \Biggr. \Biggr] .
    \end{split}
    \end{equation}
\end{lexicographic}

\subsection{Unification within Genetic Algorithm (GA)}
\label{ssec:meth:GA}

\noindent To holistically optimize modular robots, we combine
\begin{itemize}
    \item the base pose of the robot $\B \in \validBases$,
    \item the $\nModules$ modules to assemble the robot from $\assembly = [\module{1}, \ldots, \module{\nModules}], \module{i} \in \modules$, and
    \item the inverse kinematic (IK) solutions for path planning $\Jposs = [\Jposv_{\goal[1]}, \ldots, \Jposv_{\goal[\normElem{\G}]}], \Jposv_{\goal[i]} \in \IKsols_{\goal[i]}$
\end{itemize}
within a single genome.
Following \cite{Lei2024,Romiti2023}, the base pose is encoded as a vector $\basev$ added before the module encoding $\assembly$.
To implement IK optimization similar to \cite{MITSI200850,Baizid2010}, initial guesses for the IK solution $\Jpos_{\goal[i], \module{j}}$ of each goal $\goal[i]$ are added after every module $\module{j}$.

Alongside this encoding, we also define how the GA alters these genes, which is sketched in the middle of \cref{fig:overview}.
All GA parameters are summarized on the left of \cref{tab:hyperparameter}; those common to all gene combinations are population size $\popSize$, number of parents mating $\nParentsMate$, number of elites $\nElites$, number of parents to keep $\nKeepParents$, and selection pressure $\selectionPres$.
Details of each encoding and the remaining parameters are described in the following paragraphs.

\paragraph{Robot Modules}
We encode the robot assembly as a vector of $\nGenes = \nModules$ values encoding the modules to assemble from the base of the manipulator to its end effector, following~\cite{Icer2017,Klz2023OptimizingMR,Romiti2023,Lei2024}.
Each gene encodes a module or the empty module to use in that position.
Point-wise mutations can replace one module with another with a chance $\pMutate$.
To retain valid robots, the first and last element can only be replaced with another base or end effector, respectively.
Following \cite{Klz2023OptimizingMR}, the initial population is created with, on average, $\dofInit$ degrees of freedom, and the chance of mutating in the empty module is set to $\chanceEmpty$.  %

\paragraph{Robot Base as "0-th" Module}
An encoding of the robot base can be prepended as an $\nBase$-dimensional gene in front of assembly encoding, such as done by~\cite{Lei2024,Romiti2023} and resulting in $\nGenes = \nBase + \nModules$. 
This makes intuitive sense with regard to the locality desired by genomes, as the base is encoded at the same relative location in the genome and  kinematic chain.

A changed robotic base just changes $\B$ in all cost functions from \cref{ssec:hierElim}, requiring no other alterations.
Following~\cite{CoBRA}, we use the projections given for the goal tolerance in \cref{eq:projection} to represent the set of valid base poses $\validBases$ and flatten $\B$ into a vector $\basev \in \reals^\nBase$.
The base genes are mutated by adding random noise drawn from $\normalDist{\sigmaBase}{0}$.

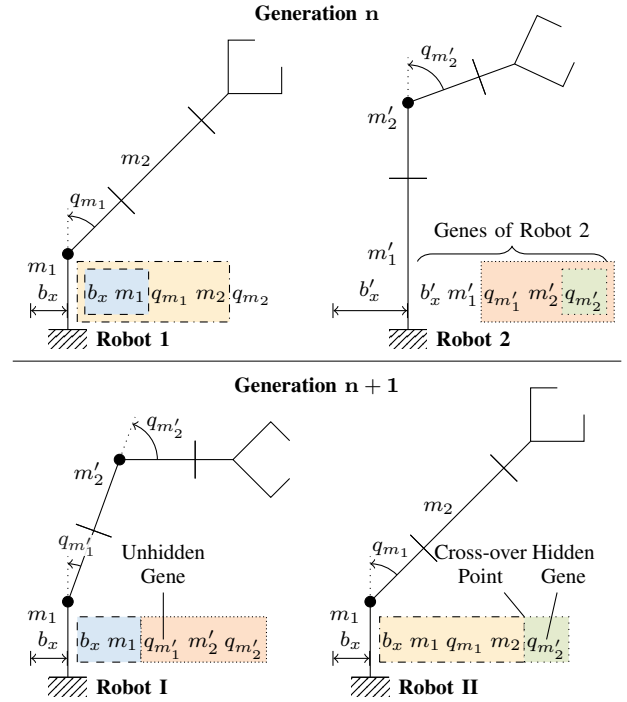
\begin{figure}[t]
    \vspace*{-0.7mm}  %
    \centering
    \usetikzlibrary{patterns,angles,quotes,fit,calc,matrix}

\newcommand{\drawJoint}[8]{

    \pgfmathsetlengthmacro{\xStart}{#1}
    \pgfmathsetlengthmacro{\yStart}{#2}
    \pgfmathsetlengthmacro{\thetaStart}{#3}
    \pgfmathsetlengthmacro{\jointAngle}{#4}    
    \pgfmathsetlengthmacro{\linkLength}{#5}
    \pgfmathsetlengthmacro{\xMiddle}{\xStart - \linkLength * sin(\thetaStart)}
    \pgfmathsetlengthmacro{\yMiddle}{\yStart + \linkLength * cos(\thetaStart)}
    \pgfmathsetlengthmacro{\connSize}{#6}

    \begin{scope}[rotate around={\thetaStart:(\xStart,\yStart)}]
        
        \draw (\xStart - \connSize/2, \yStart) -- (\xStart + \connSize/2, \yStart);
        
        \coordinate (start) at (\xStart, \yStart);
        \draw (start) --node[pos=0.8, left] {#8} (\xStart, \yStart + \linkLength) ;
    	
        \coordinate (angleStart) at (\xStart, \yStart + 1.6*\linkLength);
    	\draw[dotted] (\xStart, \yStart + \linkLength) -- (angleStart);
    \end{scope}
    \coordinate (middle) at (\xMiddle, \yMiddle);
    \draw[fill,black] (middle) circle (2pt);
    
    \pgfmathsetmacro{\thetaEnd}{\thetaStart + \jointAngle}
    
    \begin{scope}[rotate around={\thetaEnd:(\xMiddle,\yMiddle)}]
        \coordinate (end) at (\xMiddle, \yMiddle + \linkLength);
        \draw (\xMiddle, \yMiddle) -- (end);
        \draw (\xMiddle - \connSize/2, \yMiddle + \linkLength) -- (\xMiddle + \connSize/2, \yMiddle + \linkLength);
    \end{scope}

    \pic [#7, draw, ->, angle eccentricity=1.5] {angle = end--middle--angleStart};
    
    \pgfmathsetlengthmacro{\xEnd}{\xMiddle - \linkLength * sin(\thetaEnd)}
    \pgfmathsetlengthmacro{\yEnd}{\yMiddle + \linkLength * cos(\thetaEnd)}
}

\newcommand{\drawBase}[7]{
    
    \pgfmathsetlengthmacro{\xEnd}{#1}
    \pgfmathsetlengthmacro{\yEnd}{#2}
    \pgfmathsetmacro{\thetaEnd}{#3}
    
    \begin{scope}[rotate around={\thetaEnd:(\xEnd,\yEnd)}]
        
        \draw (\xEnd - #4/2, #2) -- (\xEnd + #4/2, \yEnd);

        \fill[pattern=north east lines] (\xEnd - #4/2, \yEnd) rectangle (\xEnd + #4/2, \yEnd - #4/2);
    \end{scope}
    
    \node [anchor=west] at (\xEnd + #4/2,\yEnd - #4/4) {#5};
    \draw [|<->|] (\xEnd - #6, \yEnd + #4/2) --node[pos=0.5, above] {#7} (\xEnd, \yEnd  + #4/2);
}

\newcommand{\drawLink}[6]{
    
    \begin{scope}[rotate around={#3:(#1,#2)}]
        
        \draw (#1 - #5/2, #2 + #4) -- (#1 + #5/2, #2 + #4);
        
        \draw (#1 - #5/2, #2) -- (#1 + #5/2, #2);
        
        \draw (#1, #2) --node[left] {#6} (#1, #2 + #4);
    \end{scope}
    \pgfmathsetlengthmacro{\xEnd}{#1 - #4 * sin(#3)}
    \pgfmathsetlengthmacro{\yEnd}{#2 + #4 * cos(#3)}
    \pgfmathsetmacro{\thetaEnd}{#3}
}

\newcommand{\drawEEFModule}[5]{
	
    \begin{scope}[rotate around={#3:(#1,#2)}]
        
        \draw (#1 - #5/2, #2) -- (#1 + #5/2, #2);
        
        \draw (#1, #2) -- (#1, #2 + #4);
    \end{scope}
    \pgfmathsetlengthmacro{\xEnd}{#1 - #4 * sin(#3)}
    \pgfmathsetlengthmacro{\yEnd}{#2 + #4 * cos(#3)}
    
    \path (\xEnd,\yEnd) -- ++(#3+90:2*#4) -- ([turn]-0.5*#4,0.5*#4) coordinate (one) -- ([turn]0.7071*#4,0*#4) coordinate (two) -- ([turn]0*#4,1.414*#4) coordinate (three) -- ([turn]0*#4,1.414*#4) coordinate (four) -- ([turn]0*#4,0.7071*#4) coordinate (five);
    \draw (one) -- (two) -- (three) -- (four) -- (five);
}
\def\connectorSize{5mm}
\def\rowDistance{-46mm}

\tikzset{moduleArray/.style={
  minimum width=2mm,
  inner sep=1pt,
  outer sep=0pt,
  minimum height=6mm,
  text depth=.25ex,  
  text height=1.5ex,
  execute at begin node=$,
  execute at end node=$}}
\begin{tikzpicture}[font=\footnotesize]

\begin{scope}[local bounding box=robot1box]
\drawBase{-2cm}{0cm}{0}{\connectorSize}{\textbf{Robot 1}}{5mm}{$b_x$};
\drawJoint{\xEnd}{\yEnd}{\thetaEnd}{-45}{1cm}{\connectorSize}{"$\Jpos_{m_1}$"}{$m_1$};
\drawLink{\xEnd}{\yEnd}{\thetaEnd}{1.5cm}{\connectorSize}{$m_2$};
\drawEEFModule{\xEnd}{\yEnd}{\thetaEnd}{0.5cm}{\connectorSize};
\matrix [matrix anchor=west, matrix of nodes, nodes=/tikz/moduleArray] () at (-19mm,5mm) {
	|[name=start]| b_x & |[name=co1]| m_1 & \Jpos_{m_1} & |[name=co2]| m_2 & |[name=end]| \Jpos_{m_2} \\
};
\begin{scope}[on background layer]
    \draw[dashdotted, fill=TUMBeamerYellow!20] ($(start.south west) - (1mm,1mm)$) rectangle ($(co2.north east) + (0,1mm)$);
    \draw[dashed, fill=TUMLighterBlue!20] (start.south west) rectangle (co1.north east);
\end{scope}
\end{scope}

\begin{scope}[local bounding box=robot2box]
\drawBase{2.5cm}{0cm}{0}{\connectorSize}{\textbf{Robot 2}}{10mm}{$b'_x$};
\drawLink{\xEnd}{\yEnd}{\thetaEnd}{2cm}{\connectorSize}{$m'_1$};
\drawJoint{\xEnd}{\yEnd}{\thetaEnd}{-70}{1cm}{\connectorSize}{"$\Jpos_{m'_2}$"}{$m'_2$};
\drawEEFModule{\xEnd}{\yEnd}{\thetaEnd}{0.5cm}{\connectorSize};
\matrix [matrix anchor=west, matrix of nodes, nodes=/tikz/moduleArray] () at (2.5cm,5mm) {
	|[name=start]| b'_x & m'_1 & |[name=co]| \Jpos_{m'_1} & m'_2 & |[name=end]| \Jpos_{m'_2} \\
};
\draw[decorate,decoration={brace,amplitude=2mm,mirror,raise=1mm}] (end.north east) -- (start.north west) node[midway,yshift=3mm,anchor=south]{Genes of Robot 2};

\begin{scope}[on background layer]
    \draw [densely dotted, fill=TUMOrange!20] ($(co.north west) + (0,1mm)$) rectangle ($(end.south east) + (1mm,-1mm)$);
    \draw [dotted, fill=TUMGreen!20] (end.north west) rectangle (end.south east);
\end{scope}
\end{scope}

\begin{scope}[local bounding box=robot3box]
\drawBase{-2cm}{\rowDistance}{0}{\connectorSize}{\textbf{Robot \Romannum{1}}}{5mm}{$b_x$};
\drawJoint{\xEnd}{\yEnd}{\thetaEnd}{-20}{1cm}{\connectorSize}{"$\Jpos_{m'_1}$"' fill=white, inner sep=0pt, fill opacity=0.9}{$m_1$};
\drawJoint{\xEnd}{\yEnd}{\thetaEnd}{-70}{1cm}{\connectorSize}{"$\Jpos_{m'_2}$"}{$m'_2$};
\drawEEFModule{\xEnd}{\yEnd}{\thetaEnd}{0.5cm}{\connectorSize};
\matrix [matrix anchor=west, matrix of nodes, nodes=/tikz/moduleArray] () at (-2cm,\rowDistance+5mm) {
	|[name=start]| b_x & m_1 & |[name=co]| \Jpos_{m'_1} & m'_2 & |[name=end]| \Jpos_{m'_2} \\
};
\node[above of=co] (unhiddenLabel) {\parbox{25mm}{\centering Unhidden\\ Gene}};
\draw[] (unhiddenLabel) -- (co.center);
\begin{scope}[on background layer]
    \draw [dashed, fill=TUMLighterBlue!20] (start.south west) rectangle (co.north west);
    \draw [densely dotted, fill=TUMOrange!20] (co.south west) rectangle (end.north east);
\end{scope}
\end{scope}

\begin{scope}[local bounding box=robot4box]
\drawBase{2cm}{\rowDistance}{0}{\connectorSize}{\textbf{Robot \Romannum{2}}}{5mm}{$b_x$};
\drawJoint{\xEnd}{\yEnd}{\thetaEnd}{-45}{1cm}{\connectorSize}{"$\Jpos_{m_1}$"}{$m_1$};
\drawLink{\xEnd}{\yEnd}{\thetaEnd}{1.5cm}{\connectorSize}{$m_2$};
\drawEEFModule{\xEnd}{\yEnd}{\thetaEnd}{0.5cm}{\connectorSize};
\matrix [matrix anchor=west, matrix of nodes, nodes=/tikz/moduleArray] () at (2cm,\rowDistance+5mm) {
  |[name=start]| b_x & m_1 & \Jpos_{m_1} & |[name=co]| m_2 & |[name=end]| \Jpos_{m'_2} \\
};
\node[above of=end,anchor=east] (coPointLabel) {\parbox{15mm}{\centering Cross-over Point}};
\draw[] (coPointLabel) -- (end.north west);
\node[right=-4mm of coPointLabel] (hiddenLabel) {\parbox{10mm}{\centering Hidden Gene}};
\draw[] (hiddenLabel) -- (end.center);
\begin{scope}[on background layer]
    \draw [dashdotted, fill=TUMBeamerYellow!20] (start.south west) rectangle (co.north east);
    \draw [dotted, fill=TUMGreen!20] (end.south west) rectangle (end.north east);
\end{scope}
\end{scope}

\node[fit=(robot4box) (robot3box), inner sep=2pt] (combinedLower) {};
\node[fit=(robot1box) (robot3box), inner sep=2pt] (combinedLeft) {};
\node[fit=(robot1box) (robot2box), inner sep=2pt] (combinedUpper) {};
\node[fit=(robot2box) (robot4box), inner sep=2pt] (combinedRight) {};
\draw [solid, ultra thin] (combinedUpper.south -| combinedLeft.west) -- (combinedUpper.south  -| combinedRight.east);

\node[anchor=north, fill=white, inner sep=-2pt] () at (combinedUpper.north) {\textbf{Generation} $\mathbf{n}$};
\node[anchor=north, fill=white, inner sep=6pt] () at (combinedUpper.south) {\textbf{Generation} $\mathbf{n+1}$};

\end{tikzpicture}
    \caption{
A simplified example of crossover for a single goal $\goal[1]$, which allows us to drop the first index from all IK guesses, i.e., within this figure $\Jpos_{\module{j}} = \Jpos_{\goal[1], \module{j}}$.
The different colored boxes show the genes from robots 1 and 2 that are recombined in the new robots \Romannum{1} and \Romannum{2}.
    }
    \label{fig:2DGeneExample}
\end{figure}

An example where $\nBase = 1$ is given by the robots in \cref{fig:2DGeneExample}, i.e., these robots can be positioned horizontally at different $b_x$ relative to the origin.
Therefore, $\basev = b_x$ and $\B$ is a pure translation $\Transv(\B) = \left[ b_x, 0, 0 \right]^T $.
As shown, the position $b_x$ is added in front of the genes describing each robot.

\paragraph{Selection of Inverse Kinematic (IK) Solutions}
One main challenge of optimal path planning is the selection of IK solutions \cref{ssec:relatedWork}.
For example, the Robot~\Romannum{1} in \cref{fig:2DGeneExample} could reach the same end-effector position in the ``elbow-down'' configuration, where $\Jpos_{m'_1} = \SI{-90}{\degree}$ and $\Jpos_{m'_2} \approx \SI{70}{\degree}$.
Depending on the other goals in a task, the shown or elbow-down configuration might be more cost-effective.

To jointly optimize the used IK solutions $\Jposs = [\Jposv_{\goal[1]}, \ldots, \Jposv_{\goal[\normElem{\G}]}]$ (chosen from $\IKsols_g$ for each goal $\goal{}$) and modules $\assembly$, we add random initial guesses for the IK solution of each goal $\goal[i] \in \G$ to the gene describing each module $\module{j} \in \assembly$, i.e., $\Jpos_{\goal[i], \module{j}}$.
If the module $\module{j}$ has no joint, the IK guess is \textit{hidden} \cite{Abdelkhalik2013}, i.e., it has no effect on the cost evaluation.
More implementation details are provided in App.~\ref{app:IKgene}.

The initial guesses are mutated by adding Gaussian noise with zero mean and standard deviation $\sigmaIK$ and clipping to $\realInt{0}{1}$.
To help convergence, we use \textit{Lamarckian evolution}, i.e., with a chance $\pLamarck \in \realInt{0}{1}$ we run $\nLamarck \in \naturals$ steps of the previously described IK solver (\cref{ssec:hierElim}, step~\ref{cost:ik}) and overwrite the guesses in the gene if the IK solver succeeds (adapted from \cite{MITSI200850}).

\paragraph{Crossover}
\label{ssec:meth:crossover}

As shown in \cref{fig:2DGeneExample}, the proposed gene simplifies the crossover operator, which just combines genes from any two robots in generation $n$ by choosing a single crossover point.
In this case, a crossover after the second or fourth gene creates generation $n+1$.
Especially, \cref{fig:2DGeneExample} highlights how the IK genes are (un)hidden if required by the specific module.
Robot \Romannum{1} shows a crossover where hidden information, i.e., $\Jpos_{m'_1}$, is expressed after the (passive) link module $m'_1$ is replaced by the joint module $m_1$.
Robot \Romannum{2} shows a crossover where the IK guess $\Jpos_{m'_2}$ is no longer expressed and thereby hidden, as the joint module $m'_2$ has been replaced by the link module $m_2$.

\paragraph{Special cases}
We remark that previous works optimizing modular robots, base poses, and/or robotic paths with genetic algorithms are special cases of our proposed method:
\begin{itemize}
    \item Removing the IK solutions $\Jposs$ (and replacing them with an IK solver during cost evaluation), results in \cite{Lei2024,Romiti2023};
    \item By removing the module assembly $\assembly$ (and replacing it with a fixed robotic arm) we obtain \cite{MITSI200850};
    \item After removing the assembly $\assembly$ and the base $\B$ (and assuming both are fixed) one obtains \cite{Baizid2010};
    \item By removing the IK solutions $\Jposs$ and the base $\B$, we receive \cite{Liu2020,Klz2023OptimizingMR,Icer2017,Althoff2019}.
\end{itemize}

\section{Numerical Experiments}
\label{sec:num_example}

\noindent This section covers the implementation and testing of the described optimization algorithm.
We state the software used, determine optimal hyperparameters, and compare the different optimization scopes, where optimizing the used modules $\assembly$ alone represents the baseline as previously published in \cite{Klz2023OptimizingMR,Liu2020,Althoff2019}.
Specifically, the experiments test these hypotheses: 
\begin{hypothesis}
    \item \label{hyp:better_convergence} Joint optimization of module assembly, base placement, and trajectory converges quicker to better solutions compared to independent optimizations. 
    \item \label{hyp:generalizes} Joint optimization of module assembly, base placement, and trajectory generalizes to various tasks.
    \item \label{hyp:other_costs} The optimization of cycle time benefits other common objectives.
\end{hypothesis}

\subsection{Setup}

\begin{table}[t]
    \centering
    \caption{Overview set of tasks}
    \label{tab:task_sets}
    \begin{tabular}{lcp{3.5cm}c}
        \toprule
        Set of Tasks & Task count & Description & Based on \\
        \midrule
        Simple & 100 & 3 goals, 3 cubic obstacles at random positions & \cite{Whitman2020} \\
        Hard & 100 & 5 goals, 5 cubic obstacles at random positions & \cite{Whitman2020} \\
        Real-world & 27 & 4 goals at random positions inside and outside of a 3D scanned CNC machine & \cite{Liu2020,CoBRA} \\
        Edge case & 100 & 10 different obstacle clusters surrounding one of three goals & \cite{MayerBPO} \\
        \bottomrule
    \end{tabular}
\end{table}

\noindent All considered optimization scopes are compared on robotic tasks from CoBRA~\cite{CoBRA}.
We selected four sets of tasks with various difficulties summarized in \cref{tab:task_sets} for which we optimize the cycle time $\timeJ = \timeend$.
The last set of tasks (edge case) was published in \cite{MayerBPO} and tries to make positioning and moving the robot especially difficult by placing one of the goals between obstacles.
All sets of tasks can be viewed at \href{https://redirect.cps.cit.tum.de/hmro}{redirect.cps.cit.tum.de/hmro}.

We optimize module compositions from the module set \textit{modrob-gen2}\footnote{Description:\href{https://cobra.cps.cit.tum.de/api/robots/modrob-gen2_5}{cobra.cps.cit.tum.de/api/robots/modrob-gen2\_5}, \lastChecked} without limiting availability, i.e., $n_{\mathrm{avail}}$ in \cref{eq:cost:availableModules} is set to infinity for all modules.
A detailed list of the 25 availbale modules is given in App.~\ref{app:modules} and the subset of modules available in our lab is shown in \cref{fig:moduleSet}.

The code of our optimization method is available at \href{https://gitlab.lrz.de/tum-cps/hmro}{gitlab.lrz.de/tum-cps/hmro} and is implemented in Python 3.10. Additionally, we use 
\begin{itemize}
    \item Timor-python~\cite{timor_main} to simulate each modular robot. It provides the functions defined in \cref{ssec:problem:robotmodel} and the inverse kinematics for \cref{ssec:hierElim}, step~\ref{cost:ik}f.
    \item PyGAD~\cite{gad2021pygad} as the basis for the genetic algorithm.
    \item Optuna~\cite{optuna_2019} for hyperparameter tuning of each scope.
    \item Lazy-PRM*\cref{fn:LazyPRMStar} implemented in OMPL \cite{Sucan2012} for path planning (\cref{ssec:hierElim}, step~\ref{cost:path_planning}). In contrast to previous work using RRT-C~\cite{Liu2020,Klz2023OptimizingMR}, the road map of Lazy-PRM* can be cached for each assembly $\assembly$ such that paths improve with each evaluation similar to~\cite{MayerEffPathPlan}.
    \item TOPP-RA\footnote{\href{https://pypi.org/project/toppra}{pypi.org/project/toppra}, \lastChecked}\cite{Pham2018} for trajectory generation (\cref{ssec:hierElim}, step~\ref{cost:traj_param}) as it explicitly considers the velocity limits $\Jvelvmax$ \cref{eq:kinLimit} and torque limits $\Jtorqvmax$ \cref{eq:torqueLimit} of the used joint modules.
\end{itemize}

All numerical experiments are run on the \censor{CoolMUC-4} cluster\footnote{\label{fn:coolmuc}\censor{\href{https://doku.lrz.de/coolmuc-4-1082337877.html}{doku.lrz.de/coolmuc-4-1082337877.html}}, \lastChecked} which is based on the Intel\textsuperscript{\textregistered} Xeon\textsuperscript{\textregistered} Platinum 8480+ at $\SI{2}{\giga\hertz}$.
We run each optimization on one of the $\SI{112}{}$ physical cores with about $\SI{4}{\giga\byte}$ of memory in parallel.

\subsection{Hyperparameter Tuning}
\label{ssec:numExp:HPtuning}

\noindent The hyperparameters for specific genes and the genetic algorithm in general are listed on the left of \cref{tab:hyperparameter} on the top and bottom, respectively.
Search spaces for each are given in the middle.
Even though many hyperparameters are shared between scopes, we optimize all hyperparameters for each scope individually.
For example, each scope needs to know the time $\tplan$ to do path planning for, but longer planning times are always a trade-off with spending more time, e.g., on testing more IK solutions.
We expect these trade-offs to be different for different scopes.

Each trial evaluates the optimizer on the first four tasks from each set of tasks given in \cref{tab:task_sets} -- in total 16 evaluations -- each with a fixed computation time $\ttimeout = \SI{60}{\minute}$.
Trials not having a valid solution return $\costfail = \SI{50}{\second}$, determined to be above the cost of any conceivable solution.
We give each optimization scope a budget of $\SI{400}{}$ trials to find the parameters with the best average cycle time.

To focus on promising trials, we run the 16 evaluations in parallel and report the best found $\timeJ$ values every $\SI{30}{\second}$ to the median pruner of Optuna\footnote{\label{fn:optuna} \href{https://optuna.readthedocs.io/}{optuna.readthedocs.io}, \lastChecked}.
Hyperparameters are sampled using the TPE sampler~\cite{TPE_at_NIPS} provided by Optuna\cref{fn:optuna}.
The pruner and sampler are run with the default parameters provided by Optuna 4.0.0.

\paragraph{Results}

The best hyperparameters for each optimization scope are given on the right side of \cref{tab:hyperparameter}.
These are the hyperparameters used for all of the following numerical experiments.
More details regarding the distribution of possible hyperparameter performances and the chosen cost function are given in App.~\ref{app:hps}.

\paragraph{Discussion}

As integrating IK candidates into modular robot optimization is novel in this paper, we take a closer look at the involved hyperparameters.
First, we observe that Lamarckian evolution is strongly favored and almost all valid found IK solutions from the IK solver are put back into the genome ($\pLamarck \geq \SI{0.80}{}$).
Secondly, the remembered IK guesses seem to help, as fewer iterations $\nLamarck$ are needed in comparison to $\nCplxIK$.
Lastly, the IK guesses also seem to aid path planning as $\tplan$ is lower for both cases with IK guesses.

\subsection{Comparison of Optimization Scopes}
\label{ssec:numExp:CompareScopes}

\noindent We compare the different suggested optimization scopes with the best hyperparameters listed in \cref{tab:hyperparameter}.
The scopes are compared on the remaining tasks of the suggested sets of tasks from \cref{tab:task_sets}.
Overall, this means the evaluation can run on 96 tasks from the simple, hard, and edge case datasets and on 23 of the real-world dataset.
The other parameters are the same as in \cref{ssec:numExp:HPtuning}, i.e., evaluations run for $\ttimeout = \SI{60}{\minute}$ on one core of the \censor{CoolMUC-4}\cref{fn:coolmuc}.
Each optimization is run on the five seeds given by $\naturalInt{5}$ to account for variance due to randomness in the genetic operators, IK solver, path planner, and initial populations.

\paragraph{Results}

\begin{figure*}[t]
    \vspace{-2.5mm}  %
    \centering
    \begin{tikzpicture}
  \pgfplotstableread[col sep=comma]{figures/plot_min_valid_cost.csv}\costdata  
  \pgfplotstableread[col sep=comma]{figures/plot_valid_found.csv}\validdata  
  \usepgfplotslibrary{fillbetween}
  \usepgfplotslibrary{groupplots}
  \begin{groupplot}[
    group style={
      group size=4 by 2,
      ylabels at=edge left,
      x descriptions at=edge bottom,
      horizontal sep=18pt,
      vertical sep=6pt
    },
    width=\linewidth/3.4,  
    ylabel={Best Cycle Time $\timeJ \left[ \unit{\second} \right]$\strut{}},
    y label style={at={(axis description cs:-0.12,.5)},anchor=south},
    ymin=4.8,
    ymax=9.1,
    ytick={5.0, 7.0, 9.0},
    xmode=log,
    xmin=0.4,
    xmax=72,
    xlabel={Opt. Time $\left[ \unit{\minute} \right]$},
    xtick={0.5, 1.5, 5, 20, 60},
    log ticks with fixed point,
    cycle multi list={%
      TUMDarkerBlue,TUMGreen,TUMBeamerYellow,TUMOrange\nextlist
      solid,dashed,dashed,solid  
    }%
  ]
    \nextgroupplot[title=Simple\strut{}]
    \foreach \scope in {Modules,Modules+Basexyz,Modules+IK,Modules+Basexyz+IK}{
        \addplot+ [mark=none, line width=1pt] table[x=optimization_time,y=simple.\scope.mean] {\costdata};
        \addplot+ [mark=none, name path=low] table[x=optimization_time,y=simple.\scope.low] {\costdata}; 
        \addplot+ [mark=none, name path=high] table[x=optimization_time,y=simple.\scope.high] {\costdata};     
        \addplot+ [opacity=0.1] fill between [of=low and high];
    }
    
    \nextgroupplot[
        title=Hard\strut{},
        ymin=12.5,
        ymax=18,
        ytick={13, 15, 17}]
    \addlegendimage{empty legend}  
    \foreach \scope in {Modules,Modules+Basexyz,Modules+IK,Modules+Basexyz+IK}{
        \addplot+ [mark=none, line width=1pt] table[x=optimization_time,y=hard.\scope.mean] {\costdata};
        \addplot+ [mark=none, name path=low, style=dashed] table[x=optimization_time,y=hard.\scope.low] {\costdata};
        \addplot+ [mark=none, name path=high, style=dashed] table[x=optimization_time,y=hard.\scope.high] {\costdata};
        \addplot+ [opacity=0.1] fill between [of=low and high];
    }
    
    \nextgroupplot[
        title=Real-World\strut{},
        ymin=8.5,
        ymax=13.1,
        ytick={9, 11, 13}]
    \foreach \scope/\scopeName in {Modules/Modules,Modules+Basexyz/Modules+Base,Modules+IK/Modules+IK,Modules+Basexyz+IK/Modules+Base+IK}{
        \addplot+ [mark=none, line width=1pt] table[x=optimization_time,y=realworld.\scope.mean] {\costdata};
        \addplot+ [mark=none, name path=low] table[x=optimization_time,y=realworld.\scope.low] {\costdata}; 
        \addplot+ [mark=none, name path=high] table[x=optimization_time,y=realworld.\scope.high] {\costdata};     
        \addplot+ [opacity=0.1] fill between [of=low and high];
    }
    
    \nextgroupplot[
        title=Edge Case\strut{},
        ymin=4.9,
        ymax=8.5,
        ytick={5,6.5,8}]
    \foreach \scope/\scopeName in {Modules/Modules,Modules+Basexyz/Modules+Base,Modules+IK/Modules+IK,Modules+Basexyz+IK/Modules+Base+IK}{
        \addplot+ [mark=none, line width=1pt] table[x=optimization_time,y=edgecase.\scope.mean] {\costdata};
        \addplot+ [mark=none, name path=low] table[x=optimization_time,y=edgecase.\scope.low] {\costdata}; 
        \addplot+ [mark=none, name path=high] table[x=optimization_time,y=edgecase.\scope.high] {\costdata};     
        \addplot+ [opacity=0.1] fill between [of=low and high];
    }

    \nextgroupplot[
        ymin=0,
        ymax=1,
        ytick={0, 0.2, ..., 1.0},
        ylabel={Success Rate\strut{}},
    ]
    \foreach \scope in {Modules,Modules+Basexyz,Modules+IK,Modules+Basexyz+IK}{
        \addplot+ [mark=none, line width=1pt] table[x=optimization_time,y=simple.\scope.mean] {\validdata};
        \addplot+ [mark=none, name path=low] table[x=optimization_time,y=simple.\scope.low] {\validdata}; 
        \addplot+ [mark=none, name path=high] table[x=optimization_time,y=simple.\scope.high] {\validdata};     
        \addplot+ [opacity=0.1] fill between [of=low and high];
    }
    
    \nextgroupplot[
        ymin=0,
        ymax=0.8,
        ytick={0, 0.2, ..., 0.8},
        legend style={fill opacity=0.8,
          draw opacity=1,
          text opacity=1,
          draw=none,
          font=\footnotesize,
          column sep=3pt,
          at={(1,-0.33)},
          anchor=north},
        legend cell align={left},
        legend columns=5,
    ]
    \addlegendimage{empty legend}  
    \foreach \scope in {Modules,Modules+Basexyz,Modules+IK,Modules+Basexyz+IK}{
        \addplot+ [mark=none, line width=1pt] table[x=optimization_time,y=hard.\scope.mean] {\validdata};
        \addplot+ [mark=none, name path=low, style=dashed] table[x=optimization_time,y=hard.\scope.low] {\validdata};
        \addplot+ [mark=none, name path=high, style=dashed] table[x=optimization_time,y=hard.\scope.high] {\validdata};
        \addplot+ [opacity=0.1] fill between [of=low and high];
    }
    \legend{Optimization Scope:,Modules $\assembly$,,,,Modules $\assembly$ + Base $\B$,,,,Modules $\assembly$ + IKs $\Jposs$,,,,Modules $\assembly$ + IKs $\Jposs$ + Base $\B$} 
    
    \nextgroupplot[
        ymin=0,
        ymax=1,
        ytick={0, 0.2, ..., 1.0},
    ]
    \foreach \scope/\scopeName in {Modules/Modules,Modules+Basexyz/Modules+Base,Modules+IK/Modules+IK,Modules+Basexyz+IK/Modules+Base+IK}{
        \addplot+ [mark=none, line width=1pt] table[x=optimization_time,y=realworld.\scope.mean] {\validdata};
        \addplot+ [mark=none, name path=low] table[x=optimization_time,y=realworld.\scope.low] {\validdata}; 
        \addplot+ [mark=none, name path=high] table[x=optimization_time,y=realworld.\scope.high] {\validdata};     
        \addplot+ [opacity=0.1] fill between [of=low and high];
    }
    
    \nextgroupplot[
        ymin=0,
        ymax=1,
        ytick={0, 0.2, ..., 1.0},
    ]
    \foreach \scope/\scopeName in {Modules/Modules,Modules+Basexyz/Modules+Base,Modules+IK/Modules+IK,Modules+Basexyz+IK/Modules+Base+IK}{
        \addplot+ [mark=none, line width=1pt] table[x=optimization_time,y=edgecase.\scope.mean] {\validdata};
        \addplot+ [mark=none, name path=low] table[x=optimization_time,y=edgecase.\scope.low] {\validdata}; 
        \addplot+ [mark=none, name path=high] table[x=optimization_time,y=edgecase.\scope.high] {\validdata};     
        \addplot+ [opacity=0.1] fill between [of=low and high];
    }
    \end{groupplot}
\end{tikzpicture}  %
    \caption{
        Center lines show mean cost (top) and success rate (bottom) with shaded $\qty{95}{\percent}$ confidence interval for the considered test sets (columns) and optimization scopes (color).
        Statistics are calculated over 23 (real world) or 96 (all other) tasks and five seeds.
    }
    \label{fig:convergence}
\end{figure*}
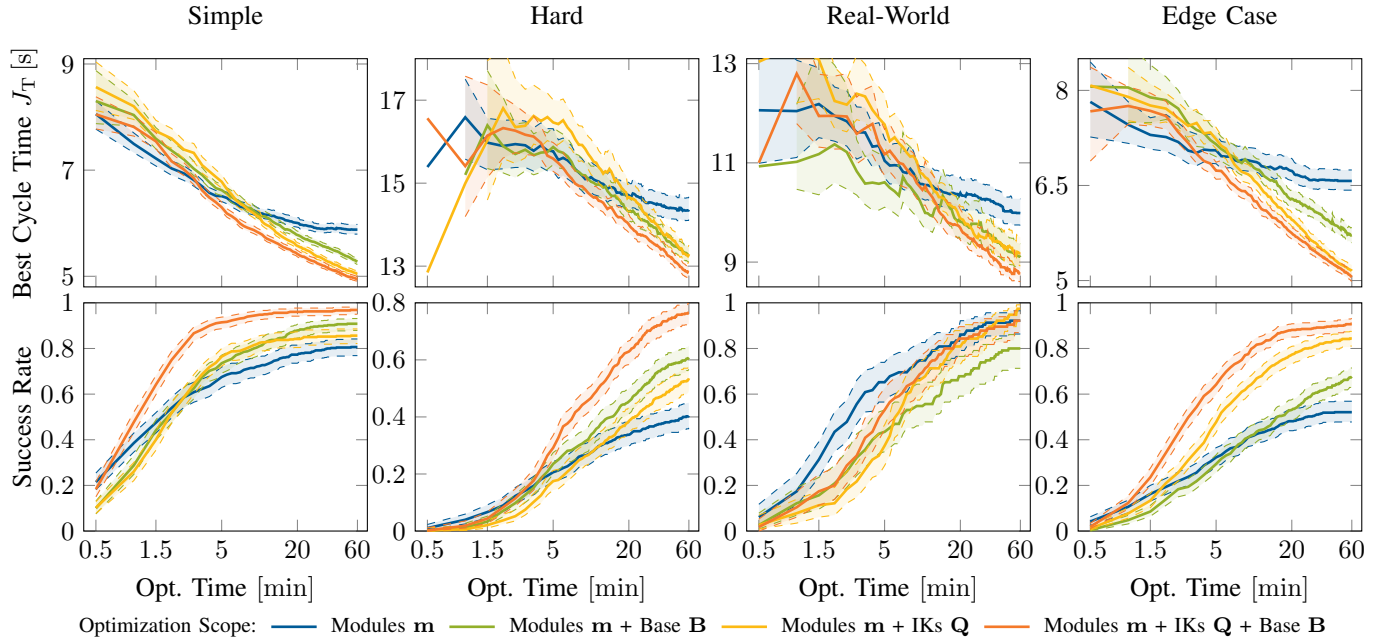

\begin{table}[t]
    \vspace{-1.7mm}  %
    \centering
    \caption{
    Confidence intervals of relative performance at $\ttimeout$ vs. only optimizing the module assembly $\assembly$. 
    $\uparrow, \downarrow$ indicate the direction of better values and \textbf{bold} the best values per set of tasks.}
    \label{tab:performance_gain}
    \begin{tabular}{lcR@{, }LR@{, }L}  %
    \toprule
        Set of Tasks & Scope & \multicolumn{2}{c}{$\downarrow$ Cycle Time $\unit{\percent}$} & \multicolumn{2}{c}{$\uparrow$ Success Rate $\unit{\percent}$} \\
    \midrule
        \multirow{3}{*}{Simple} & $\scopeMB$ & 
        [ \SI{-12.0}{} & \SI{-8.7}{} ] & 
        [ \SI{7.5}{} & \SI{18.4}{} ] \\
        & $\scopeMQ$ & 
        [ \SI{-15.8}{} & \SI{-12.8}{} ] & 
        [ \SI{3.4}{} & \SI{9.4}{} ] \\
        & $\scopeMBQ$ & 
        [ \mathbf{\SI{-17.4}{}} & \mathbf{\SI{-14.6}{}} ] & 
        [ \mathbf{\SI{15.7}{}} & \mathbf{\SI{25.9}{}} ] \\
    \midrule
        \multirow{3}{*}{Hard} & $\scopeMB$ & 
        [ \SI{-9.7}{} & \SI{-5.3}{} ] & 
        [ \SI{33.0}{} & \SI{71.1}{} ] \\
        & $\scopeMQ$ & 
        [ \SI{-9.8}{} & \SI{-5.3}{} ] & 
        [ \SI{17.2}{} & \SI{47.9}{} ] \\
        & $\scopeMBQ$ & 
        [ \mathbf{\SI{-12.6}{}} & \mathbf{\SI{-8.4}{}} ] & 
        [ \mathbf{\SI{70.4}{}} & \mathbf{\SI{112.3}{}} ] \\  %
    \midrule
        \multirow{3}{*}{Real-World} & $\scopeMB$ & 
        [ \SI{-11.9}{} & \SI{-5.9}{} ] & 
        [ \SI{-22.0}{} & \SI{-4.9}{} ]\\
        & $\scopeMQ$ & 
        [ \SI{-11.4}{} & \SI{-4.2}{} ] & 
        [ \mathbf{\SI{0.9}{}} & \mathbf{\SI{12.9}{}} ]\\
        & $\scopeMBQ$ & 
        [ \mathbf{\SI{-14.7}{}} & \mathbf{\SI{-8.2}{}} ] & 
        [ \SI{-6.4}{} & \SI{6.9}{} ]\\
    \midrule
        \multirow{3}{*}{Edge Case} & $\scopeMB$ & 
        [ \SI{-15.8}{} & \SI{-10.5}{} ] & 
        [ \SI{19.0}{} & \SI{41.1}{} ]\\
        & $\scopeMQ$ & 
        [ \SI{-23.9}{} & \SI{-19.3}{} ] & 
        [ \SI{48.5}{} & \SI{76.0}{} ]\\
        & $\scopeMBQ$ & 
        [ \mathbf{\SI{-25.2}{}} & \mathbf{\SI{-21.1}{}} ] & 
        [ \mathbf{\SI{60.4}{}} & \mathbf{\SI{89.2}{}} ]\\
    \bottomrule
    \end{tabular}
\end{table}

Each optimization run logs all tested individuals and the value of the evaluated cost functions\footnote{Logs and valid solutions available at \href{https://redirect.cps.cit.tum.de/hmro}{redirect.cps.cit.tum.de/hmro} $\rightarrow$ Data. \lastChecked}.
Based on these logs, we can plot the average shortest cycle time $\timeJ$ and fraction of solved tasks up to a certain optimization time, as shown in \cref{fig:convergence}.
Additionally, we calculate $\SI{95}{\percent}$ confidence intervals (via bootstrapping over the difference of means) for the percentage change in cycle time $\timeJ$ and success rate at the end of optimization after $\ttimeout = \SI{60}{\minute}$ versus the baseline of only optimizing the used modules $\assembly$.
These values are given in \cref{tab:performance_gain}.

\begin{table}[t]
    \vspace{-1.3mm}  %
    \caption{Correlation of cycle time $\timeJ$ with other costs}
    \label{tab:other_costs}
    \sisetup{table-format=-1.3,
             round-mode=places,
             round-precision=3,
             table-column-width=5em}
    \centering
    \begin{tabular}{lR@{ }S@{, }S@{ }L}  %
    \toprule
    Other Cost & \multicolumn{4}{C}{\text{\makebox[0pt]{Pearson Corr. Coef. 95\% CI}}} \\
    \midrule
Traj. length joint space & [ & 0.8921334856493272  &  0.8939488061699882 & ] \\
Mechanical energy & [ & 0.5718232074575589  &  0.5778263792745515 & ] \\
Number of joints & [ & 0.07570070330076122  &  0.08460878469366738 & ] \\
Number of modules & [ & -0.08482994264755277  &  -0.07592217979590896 & ] \\
Robot mass & [ & -0.08731572757667484  &  -0.0784116055469403 & ] \\
    \bottomrule
    \end{tabular}
\end{table}

In \cref{tab:other_costs}, we correlate the main objective cycle time $\timeJ$ with other often secondary costs, such as energy consumption or robot complexity.
As specific module costs are unavailable, we consider the robot mass, number of modules, and number of joints as proxies.
We use the Pearson correlation coefficient, which tests for linear correlation between two datasets, with $1$ indicating a perfect linear relationship, $-1$ a negative correlation, and $0$ no correlation.
For each cost we give the $\SI{95}{\percent}$ confidence interval of the correlation coefficient.

\paragraph{Discussion}

Our primary focus lies on \cref{fig:convergence}, as convergence and success rate over time are key performance indicators for the any-time optimization algorithm we analyze.
\cref{tab:performance_gain} is used to compare final performance, i.e., quantify the differences in \cref{fig:convergence} at $\ttimeout = \SI{60}{\minute}$.
Overall, the simple and hard tasks set form the boundary of possible task complexities.
The Friedman test for differences confirms that there are statistically signifcant differences between optimization scopes (see \cref{tab:friedman}).

The real-world tasks based on 3D scans seem to be only marginally more complex than the simple tasks.
These tasks struggle to differentiate optimization scopes besides $\scopeMB$, which was suggested in \cite{Lei2024,Romiti2023}.
This scope lags behind the other scopes significantly over most of the optimization time and is the only case of significantly deteriorated performance in \cref{tab:performance_gain}.
This set of tasks also experiences bigger confidence intervals (due to fewer tasks included), making judgments about significant differences hard.

The next set of tasks (edge case) strongly favors global IK optimization.
We suggest that path planning and especially finding any valid IK is hard in these tasks, so accumulating IK solutions over time and sharing them between similar assemblies seems to help a lot for convergence and minimizing mean cycle time.
Still, the edge case tasks also care about the location of the robot base, i.e., removing the base from the optimization scope leads to significantly lower success rates.

Lastly, the hard set of tasks favors optimizing the robot base over optimizing the IK solutions jointly with modules. 
This is probably due to the increased number of obstacles that make it hard to move any robot at the suggested base pose.

Overall, module optimization alone -- the previous state of the art in \cite{Klz2023OptimizingMR,Liu2020,Althoff2019,Icer2017} -- has significantly higher average costs in all sets of tasks and lower success rates in three of four sets of tasks, which is also confirmed via the pair-wise post-hock analysis comparing optimization scopes in \cref{tab:post_hoc_pairwise}.
At most optimization times, the novel and biggest optimization scope $\scopeMBQ$ results in significant decreases in cost and increases in the fraction of tasks solved (see its mean outside any other confidence interval in \cref{fig:convergence}), supporting hypothesis \ref{hyp:better_convergence} and superseding previous work on the joint optimization of assemblies and base poses ($\scopeMB$) in \cite{Lei2024,Romiti2023}.
In all sets of tasks, the biggest optimization scope $\scopeMBQ$ minimizes cost significantly faster and in three of four sets of tasks has a significantly higher rate of success (bold numbers in \cref{tab:performance_gain}), supporting hypothesis \ref{hyp:generalizes}.

Regarding other costs of interest in \cref{tab:other_costs}, we find that our main objective of minimizing cycle time has an inconsistent effect on robot complexity, increasing the number of modules and mass but decreasing the number of joints.
Other metrics, such as the trajectory length in joint space and mechanical energy consumed by the robot, are minimized alongside the cycle time, confirming \ref{hyp:other_costs}.

\section{Real-world Validation}
\label{sec:real_world_example}

\noindent Our real-world experiments validate that
\begin{hypothesis}
\item \label{val:samePerformance} the optimized robots achieve comparable performance in the real world, and 
\item \label{val:smallAdaptation} adapting to the sim-to-real gap does not increase programming time much.
\end{hypothesis}
The major steps of our experiments are:
\begin{enumerate}
    \item Scan the robot working area with the 3D Scanner App\footnote{\href{https://3dscannerapp.com/}{3dscannerapp.com}, \lastChecked} on an $\SI{12.9}{\inch}$ iPad Pro $4^{\mathrm{th}}$ Gen, e.g., task in \cref{fig:overview}.
    \item Define the robotic task in the generated 3D scan, e.g., annotations on top of the task in \cref{fig:overview}.
    \item Optimize the modular robot following \cref{sec:implementation} and using the best hyperparameters from \cref{ssec:numExp:HPtuning}.
    \item Deploy the optimized robot (shown in \cref{fig:task_solutions}) to measure its performance.
    \item (If required) Adapt the robot base position or trajectory to resolve collisions or other constraints.
\end{enumerate}

\subsection{Required Adaptations}

\noindent We validate our approach with two tasks using the modules produced by \censor{RobCo\cref{fn:robco}} as the physical implementation of modrob-gen2.
All robots are programmed with the graphical interface \censor{RobFlow}, which can use the via points generated by our path planning cost function of Step~\ref{cost:path_planning} in  \cref{ssec:hierElim} as desired point-to-point movements.
The tasks are named \textit{\textbf{A}round} (shown in \cref{fig:task_solutions}, top) and \textit{\textbf{B}etween} (shown in \cref{fig:task_solutions}, bottom), and the remaining assemblies are shown on the bottom of \cref{fig:overview}.

We expected multiple sources of discrepancies between our simulation and real-world experiments, which we list in App.~\ref{app:sim_real_gap}.
We especially highlight that we have a limited number of available modules, which we enforce by setting $n_{\mathrm{avail}}$ in \cref{eq:cost:availableModules}.
Still, $\SI{28766094}{}$ combinations with $\leq \SI{12}{}$ modules remain.  %

Additionally, we increased the safety margin of the collision checker used for path planning in step~\ref{cost:path_planning} and \ref{cost:solution} from $\SI{1}{\centi\meter}$ to $\SI{3}{\centi\meter}$ to account for the sources of uncertainty in App.~\ref{app:sim_real_gap}.

\subsection{Results}

\begin{table}[t]  %
    \vspace{-1.7mm}  %
    \caption{Real-world Validation (S = small and fixable collisions)}
    \label{tab:validation_res}
    \centering
    \begin{tabular}{llccccc}
    \toprule
        Task & Seed & 1 & 2 & 3 & 4 & 5\\
    \midrule
        \multirow{4}{*}{\rotatebox[origin=c]{90}{
            \parbox{1.5cm}{\centering\textbf{A}round \\
            \cref{fig:task_solutions} (A)}}
        } & Sim. cyc. time $\timeJ$ $[\unit{\second}]$ & $\SI{1.42}{}$ & $\SI{1.89}{}$ & $\SI{1.86}{}$ & $\SI{1.64}{}$ & $\SI{2.09}{}$\\
        & Real-world $\timeJ$ in $[\unit{\second}]$ & $\SI{2.64}{}$ & $\SI{1.45}{}$ & $\SI{1.45}{}$ & $\SI{2.71}{}$ & $\SI{1.65}{}$ \\
        & Collided & \xmark & S & \xmark & S & S \\
        & Task solved & \cmark & \cmark & \cmark & \cmark & (\cmark) \\
    \midrule
        \multirow{4}{*}{\rotatebox[origin=c]{90}{
            \parbox{1.5cm}{\centering\textbf{B}etween \\
            \cref{fig:task_solutions} (B)}}
        } & Sim. cyc. time $\timeJ$ $[\unit{\second}]$ & $\SI{1.97}{}$ & $\SI{2.18}{}$ & $\SI{1.27}{}$ & $\SI{1.65}{}$ & $\SI{1.53}{}$ \\
        & Real-world $\timeJ$ in $[\unit{\second}]$ & $\SI{2.95}{}$ & $\SI{2.99}{}$ & $\SI{2.35}{}$ & $\SI{3.87}{}$ & $\SI{2.56}{}$ \\
        & Collided & S & S & \xmark & \xmark & S \\
        & Task solved & \cmark & \cmark & \cmark & \cmark & \cmark \\
    \bottomrule
    \end{tabular}
\end{table}

\begin{figure*}
    \centering
    \begin{tikzpicture}
    \matrix[column sep=4pt, row sep=4pt] {
        \node[rotate=90] () {\textbf{A}round, Seed 4}; &
        \node[inner sep=0] (imageA1) {
            \includegraphics[width=0.45\columnwidth, trim={1cm 2cm 1cm 6cm}, clip]{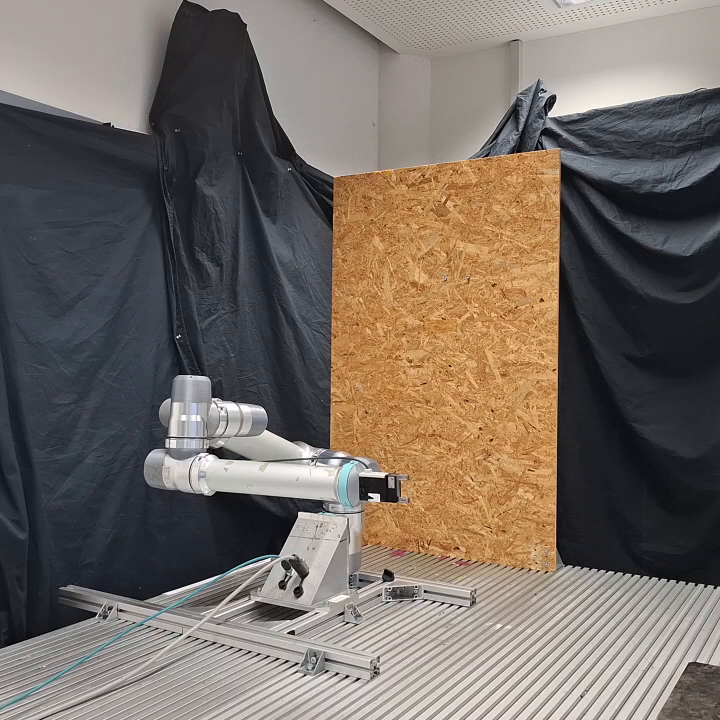}}; &
        \node[inner sep=0] (imageA2) {
            \includegraphics[width=0.45\columnwidth, trim={1cm 2cm 1cm 6cm}, clip]{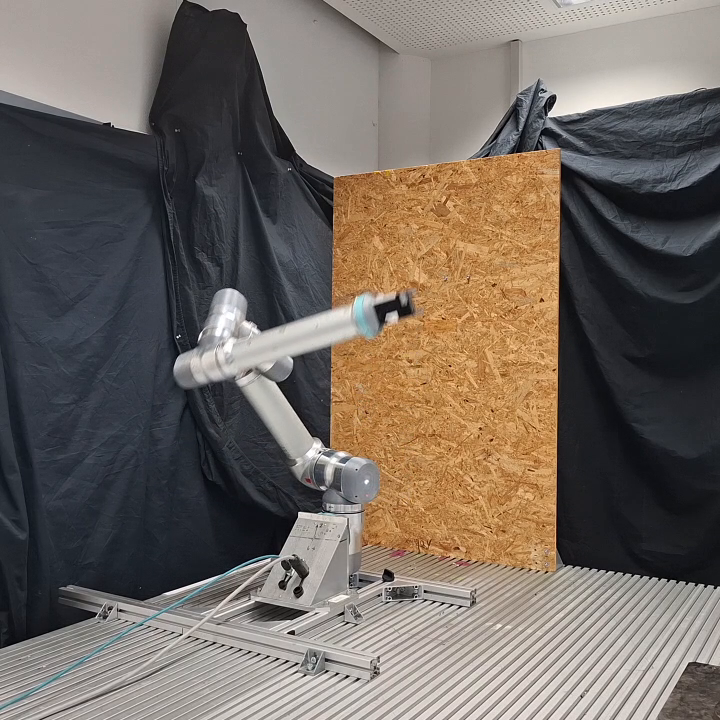}}; &
        \node[inner sep=0] (imageA3) {    
            \includegraphics[width=0.45\columnwidth, trim={1cm 2cm 1cm 6cm}, clip]{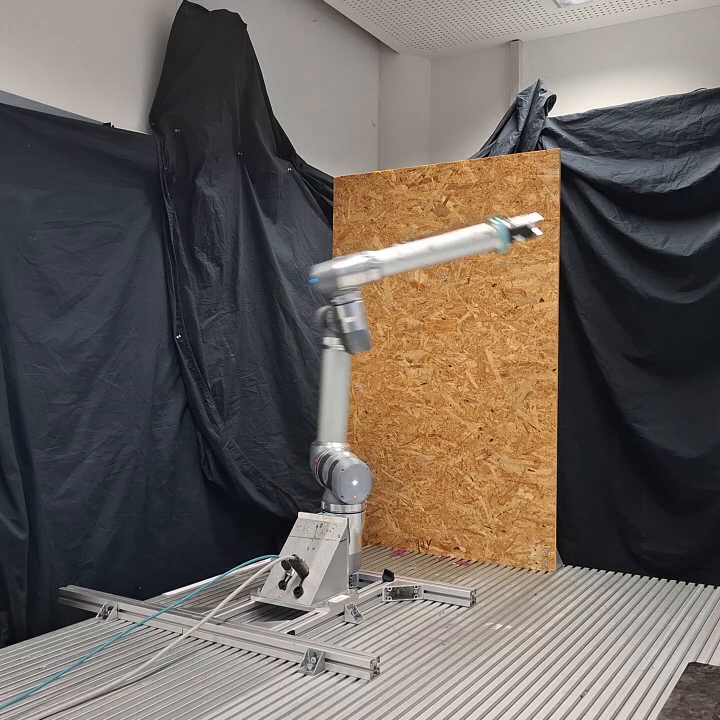}}; &
        \node[inner sep=0] (imageA4) {    
            \includegraphics[width=0.45\columnwidth, trim={1cm 2cm 1cm 6cm}, clip]{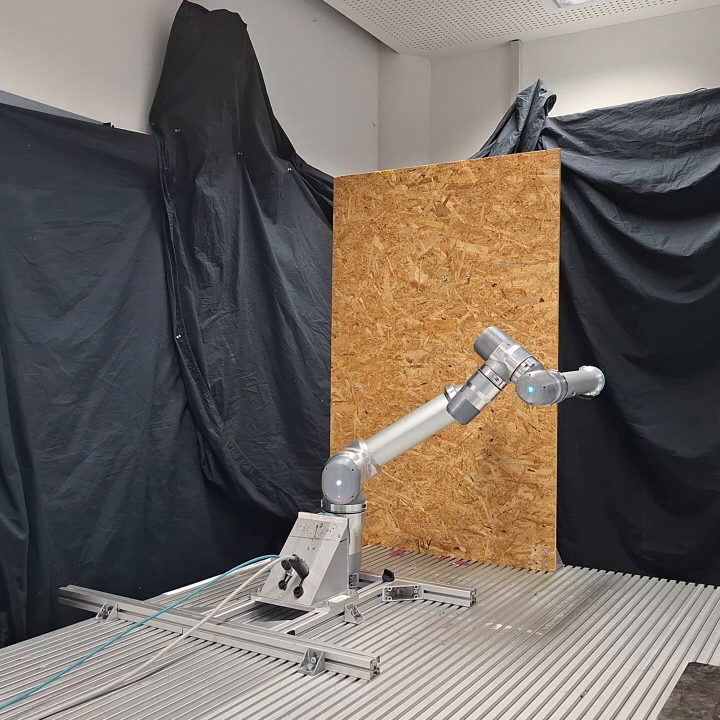}}; \\
        \node[rotate=90] () {\textbf{B}etween, Seed 3}; &
        \node[inner sep=0] (imageB1) {    
            \includegraphics[width=0.45\columnwidth, trim={1cm 4cm 1cm 4cm}, clip]{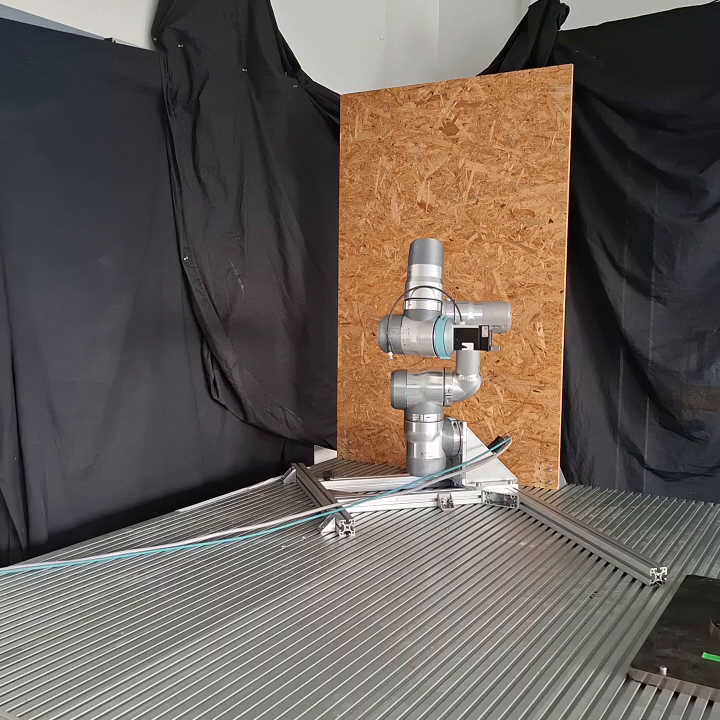}}; &
        \node[inner sep=0] (imageB2) {    
            \includegraphics[width=0.45\columnwidth, trim={1cm 4cm 1cm 4cm}, clip]{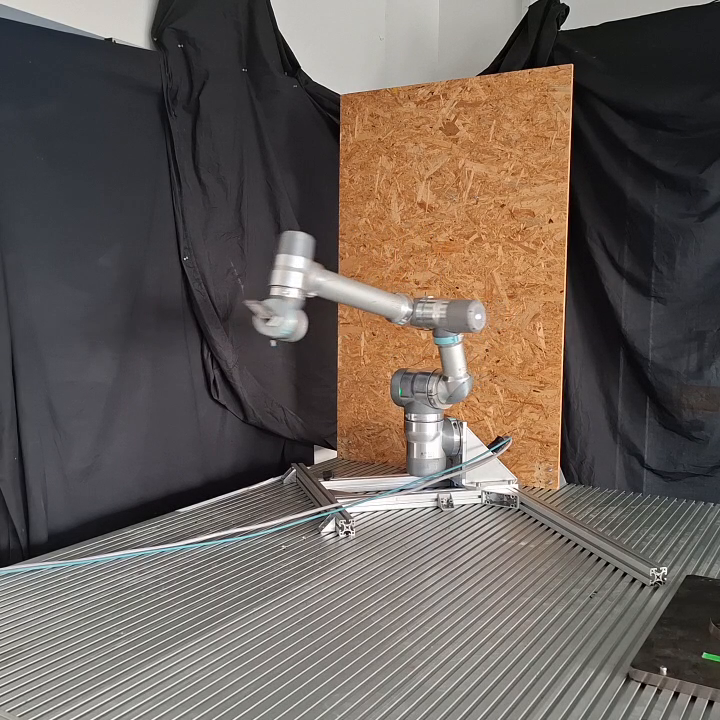}}; &
        \node[inner sep=0] (imageB3) {    
            \includegraphics[width=0.45\columnwidth, trim={1cm 4cm 1cm 4cm}, clip]{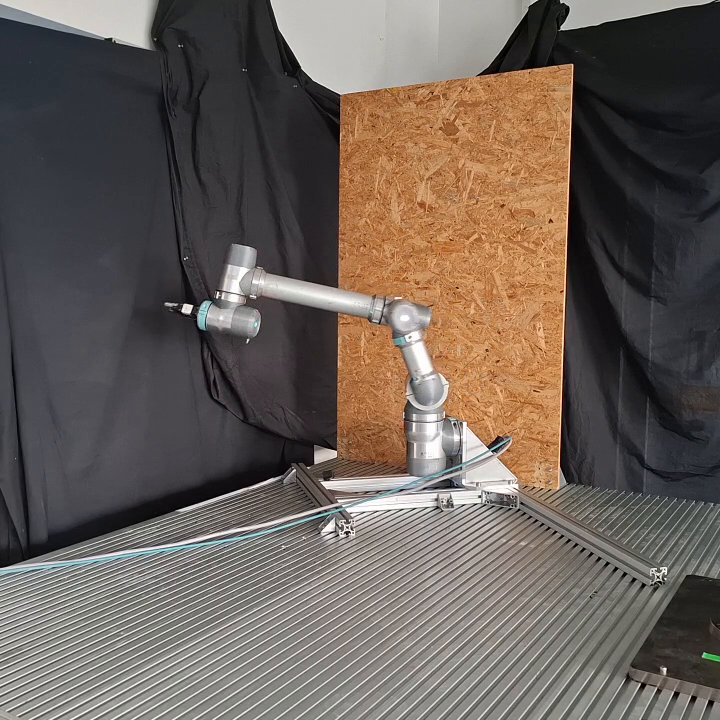}}; &
        \node[inner sep=0] (imageB4) {    
            \includegraphics[width=0.45\columnwidth, trim={1cm 4cm 1cm 4cm}, clip]{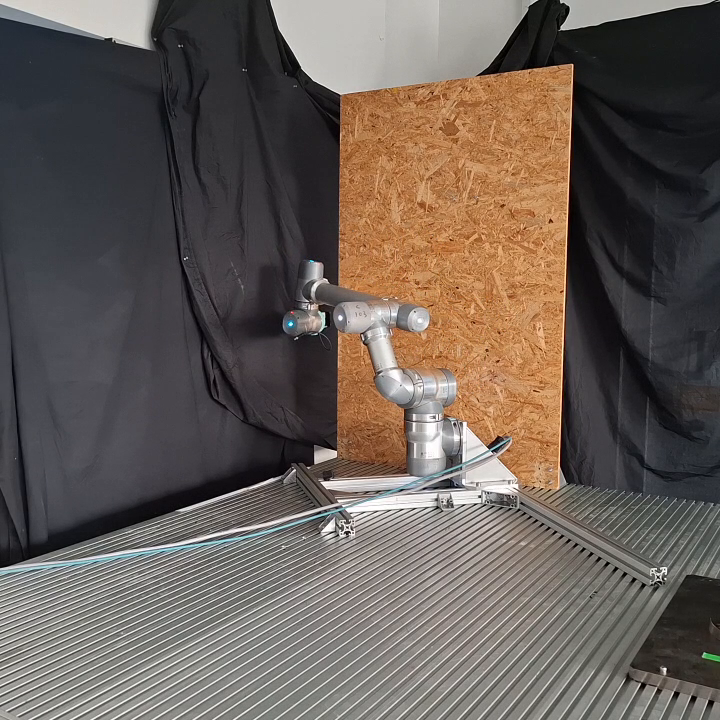}}; \\
    };

    \draw[->,thick] ($(imageB1.south west) - (0, 12pt)$) -- ($(imageB4.south east) - (0, 12pt)$);
    \node[below=12pt of imageB1, anchor=center,label=below:$0$] () {$|$};
    \node[below=12pt of imageB2, anchor=center,label=below:$t_1$] () {$|$};
    \node[below=12pt of imageB3, anchor=center,label=below:$t_2$] () {$|$};
    \node[below=12pt of imageB4, anchor=center,label=below:$\timeJ$] () {$|$};

    \begin{scope}[
        inner sep=2pt,
        font={\footnotesize},
        line width=1pt]
        \begin{scope}[
            x={($(imageA1.south east) - (imageA1.south west)$)},
            y={($(imageA1.north west) - (imageA1.south west)$)},
            shift={(imageA1.south west)},
        ]
            
            \node (CoScenterX) [white, inner sep=0, outer sep=0] at (0.37,0.1888) {};
            \node (CoScenterY) [white, inner sep=0, outer sep=0] at (0.5,0.215) {};
            \node (CoScenterZ) [white, inner sep=0, outer sep=0] at (0.45,0.45) {};
            \node [NiceRed,rectangle,fill=lightgray] (xend) at (0.17,0.11) {$x$};
            \node [NiceGreen,rectangle,fill=lightgray] (yend) at (0.65,0.2) {$y$};
            \node [NiceBlue,rectangle,fill=lightgray] (zend) at (0.45,0.7) {$z$};
            \draw [->,NiceRed] (CoScenterX) to (xend);
            \draw [->,NiceGreen] (CoScenterY) to (yend);
            \draw [->,NiceBlue] (CoScenterZ) to (zend);
        
            \node [circle,fill=blue,minimum size=7mm,opacity=0.3] (goal1) at (0.6,0.35) {};
            \node [NiceBlue,fill=lightgray] (g1text) at (0.77, 0.33) {Goal 1};
            
        \end{scope}

        \begin{scope}[
            x={($(imageA2.south east) - (imageA2.south west)$)},
            y={($(imageA2.north west) - (imageA2.south west)$)},
            shift={(imageA2.south west)},inner sep=2pt
        ]
            \node [fill=lightgray] (D116label) at (0.7, 0.33) {D116};
            \draw[white] (D116label) to (0.52,0.35);
            \draw[white] (D116label) to (0.5,0.25);
    
            \node [fill=lightgray] (D116350label) at (0.7, 0.5) {L116-350};
            \draw[white] (D116350label) to (0.4,0.5);
    
            \node [fill=lightgray] (I86350label) at (0.4, 0.85) {I86-350};
            \draw[white] (I86350label) to (0.4,0.68);
            
            \node [fill=lightgray] (D86label) at (0.25, 0.4) {D86};
            \draw[white] (D86label) to (0.25,0.56);
            \draw[white] (D86label) to (0.28,0.68);
            \draw[white] (D86label) to (0.37,0.6);
            
            \node [black,fill=lightgray] (sledtext) at (0.8, 0.1) {Sled};
            \draw[white] (sledtext) to (0.65,0.13);
        \end{scope}
        
        \begin{scope}[
            x={($(imageA3.south east) - (imageA3.south west)$)},
            y={($(imageA3.north west) - (imageA3.south west)$)},
            shift={(imageA3.south west)}]
            
            \node [fill=lightgray] (D86label) at (0.25, 0.6) {D86};
            \draw[white] (D86label) to (0.44,0.7);
            
            \node [fill=lightgray] (D86label) at (0.75, 0.7) {Gripper};
            \draw[white] (D86label) to (0.75,0.86);
            
            \node [fill=lightgray,anchor=south] (baselabel) at (0.4, 0.02) {
                \footnotesize Base at $(\SI{13}{\centi\meter}, \SI{5}{\centi\meter})$};
            \draw[white] (baselabel) to (0.43,0.25);
        \end{scope}

        \begin{scope}[
            x={($(imageA4.south east) - (imageA4.south west)$)},
            y={($(imageA4.north west) - (imageA4.south west)$)},
            shift={(imageA4.south west)}]
    
            \node [circle,fill=blue,minimum size=5mm,opacity=0.3] (goal2) at (0.87,0.57) {};
            \node [NiceBlue,fill=lightgray] (g2text) at (0.85, 0.4) {Goal 2};
        \end{scope}

        \begin{scope}[
            x={($(imageB1.south east) - (imageB1.south west)$)},
            y={($(imageB1.north west) - (imageB1.south west)$)},
            shift={(imageB1.south west)}
        ]
        
            \node (CoScenter) [white, inner sep=0, outer sep=0] at (0.45,0.22) {};
            \node [NiceRed,rectangle,fill=lightgray] (xend) at (0.27,0.15) {$x$};
            \node [NiceGreen,rectangle,fill=lightgray] (yend) at (0.65,0.2) {$y$};
            \node [NiceBlue,rectangle,fill=lightgray] (zend) at (0.45,0.6) {$z$};
            \draw [->,NiceRed] (CoScenter) to (xend);
            \draw [->,NiceGreen] (CoScenter) to (yend);
            \draw [->,NiceBlue] (CoScenter) to (zend);
        
            \node [circle,fill=blue,minimum size=7mm,opacity=0.3] (goal1) at (0.75,0.55) {};
            \node [NiceBlue,fill=lightgray] (g1text) at (0.83, 0.43) {Goal 1};
        
        \end{scope}

        \begin{scope}[
            x={($(imageB2.south east) - (imageB2.south west)$)},
            y={($(imageB2.north west) - (imageB2.south west)$)},
            shift={(imageB2.south west)}
        ]
            
            \node [fill=lightgray] (D116label) at (0.4, 0.35) {D116};
            \draw[white] (D116label) to (0.57,0.42);
            \draw[white] (D116label) to (0.6,0.3);
            
            \node [fill=lightgray,anchor=south] (baselabel) at (0.6, 0.02) {
                \footnotesize Base at $(\SI{-28}{\centi\meter}, \SI{20}{\centi\meter})$};
            \draw[white] (baselabel) to (0.7,0.27);

            \node [fill=lightgray] (L86165Label) at (0.85, 0.5) {L86-165};
            \draw[white] (L86165Label) to (0.65,0.5);
            
            \node [fill=lightgray] (D86label) at (0.78, 0.75) {D86};
            \draw[white] (D86label) to (0.58,0.6);
            \draw[white] (D86label) to (0.65,0.6);
            
            \node [fill=lightgray] (I86350label) at (0.55, 0.8) {I86-350};
            \draw[white] (I86350label) to (0.5,0.65);
            
            \node [fill=lightgray] (D86label) at (0.15, 0.6) {Gripper};
            \draw[white] (D86label) to (0.35,0.58);

        \end{scope}

        \begin{scope}[
            x={($(imageB3.south east) - (imageB3.south west)$)},
            y={($(imageB3.north west) - (imageB3.south west)$)},
            shift={(imageB3.south west)}
        ]
            \node [fill=lightgray] (D86label) at (0.5, 0.75) {D86};
            \draw[white] (D86label) to (0.32,0.7);
            \draw[white] (D86label) to (0.3,0.58);
            
            \node [fill=lightgray,anchor=south] (baselabel) at (0.6, 0.02) {
                \footnotesize Rotated $\SI{136}{\degree}$ around \color{NiceBlue}$z$};
            \draw[white] (baselabel) to (0.7,0.27);
            
        \end{scope}

        \begin{scope}[
            x={($(imageB4.south east) - (imageB4.south west)$)},
            y={($(imageB4.north west) - (imageB4.south west)$)},
            shift={(imageB4.south west)}
        ]
            
            \node [circle,fill=blue,minimum size=5mm,opacity=0.2] (goal2) at (0.45,0.55) {};
            \node [NiceBlue,fill=lightgray] (g2text) at (0.35, 0.43) {Goal 2};
        \end{scope}  %
    \end{scope}  %
    
    \end{tikzpicture}
    \caption{
        The best robots found for task \textit{\textbf{A}round}, seed 4 (top) and \textit{\textbf{B}etween}, seed 3 (bottom) moving between goal 1 and 2. 
        Both videos are available\cref{fn:paper_webpage}.}
    \label{fig:task_solutions}
\end{figure*}
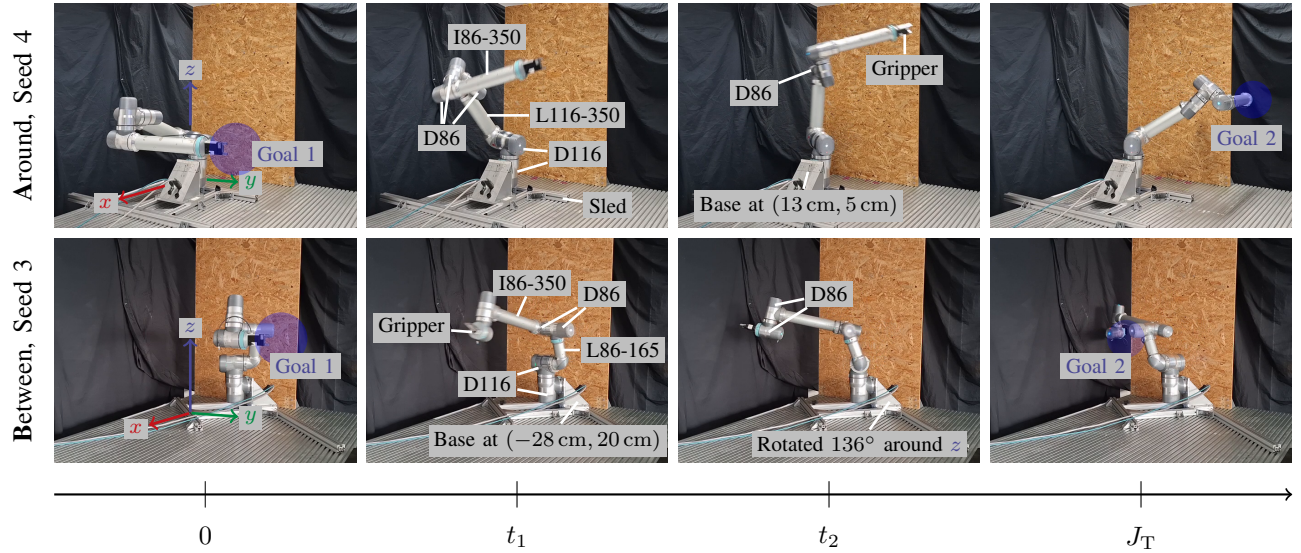

\noindent In \cref{tab:validation_res}, we note the primary observation about the tested optimized robots.
Each robot was manually assembled on average in $\SI{35}{\minute}$, including the disassembly of the previous robot.
Programming the robot to solve the task with the planned path and recording its movement took on average $\SI{19}{\minute}$.

Overall, the task was satisfied in nine out of ten cases, and all goals were reached without collisions.
We found similar results when stress-testing the optimized solutions, showing that they can be adapted to $\SI{95}{\percent}$ of random goal shifts (see \cref{app:stressTest}).

Looking closer at the individual experiments, the only exception is seed 5 for task A, which was not reliably executable due to a joint being close to its rated torque for most of the optimized trajectory.
Otherwise, in six out of ten cases, adaptations to the optimization results were required due to collisions, as noted in \cref{tab:validation_res}.
All changes were minor; joint positions were manually altered in four cases, and the robot base was moved in two cases.
If the suggested trajectory collided with the environment, programming time increased on average by $\SI{8}{\minute}$ or less than $\SI{50}{\percent}$, supporting \ref{val:smallAdaptation}.
In two cases, the robot only collided with its unmodelled sled and the holder for the black curtains, which were added for filming the robot.

On average, the cycle time is $\SI{0.20}{\second}$ (A) and $\SI{1.22}{\second}$ slower (B) than in simulation or roughly $\SI{10}{\percent}$ and $\SI{42}{\percent}$, respectively, further supporting \ref{val:samePerformance}.
Task B, in particular, often had more PTP points to handle the narrow passage the robot had to move into to reach goal 2 (see \cref{fig:task_solutions}, B).
The required stop at each PTP point was partially mitigated by allowing path smoothing, i.e., a blending at the intermediate joint space positions.

\section{Conclusions} \label{sec:conclusions}

\noindent For the first time, this paper applied holistic optimization of modular robots, i.e., jointly optimizing the module selection, base position, and executed trajectory, to solve point-to-point movements -- the most common industrial task.
In numerical experiments with over $\SI{300}{}$ tasks, we showed that this holistic optimization dominates the previous methods at most optimization budgets returning lower-cost solutions with a higher chance of successfully solving the task.
Especially for difficult tasks with cluttered environments or more goals to reach, our enlarged optimization scope increased the success rate by $\SI{60}{\percent}$ to $\SI{112}{\percent}$ relative to the state of the art.

Also for the first time, we test the real-world applicability of these optimization results by deploying two distinct tasks and five trials within our lab.
We show that the optimization results are often directly successful on the real robot (four out of ten cases), and manual fixes to avoid collisions made almost all optimization results usable (nine out of ten cases) with little extra effort.
In summary, this extended evaluation gives practitioners more confidence in trusting optimized modular robots and guides the community to focus their efforts on aspects that still struggle to transfer to the real world.

\section*{Acknowledgements}
\noindent We acknowledge the \xblackout{Leibniz Supercomputing Centre} for funding this project by providing computing time on its Linux cluster.
We thank \xblackout{the RobCo GmbH and, specifically, Paul Maroldt} for their help in understanding the API of their robot.
Additionally, we thank \xblackout{Jonathan K\"ulz for his cooperation on Timor Python and MCS}, as well as our students \xblackout{Lukas Hornik and Daniel Ostermeier} for their work on 3D scanning.

\bibliography{mmayerReferences_clean}
\bibliographystyle{IEEEtranWithoutAddress}

\appendix

\subsection{Implementation of the IK Gene}
\label{app:IKgene}

\noindent Another challenge besides the change from static link to joint module is the handling of the various joint limits coming from different joint modules $\module{j}$.
That is why the IK gene (usage described in \cref{ssec:meth:GA}) does not encode the absolute joint position $\Jpos_{\goal[i], \module{j}}$ required to reach the goal $\goal[i]$, but rather the joint position relative to the limits $\Jposvmin, \Jposvmax$.
Formally, we add the IK gene $\rho_{\goal[i], \module{j}} \in \realInt{0}{1}$ for each goal $\goal[i]$ after each gene encoding a module $\module{j}$.
If the module $\module{j}$ contains a joint with limits $\Jposvmin[j],\Jposvmax[j]$, the gene $\rho_{\goal[i], \module{j}}$ is scaled to the joint limits, i.e., the IK solution for goal $\goal[i]$ is:

\begin{equation}
    \Jpos_{\goal[i], \module{j}} = (1-\rho_{\goal[i], \module{j}}) \Jposvmin[j] + \rho_{\goal[i], \module{j}} \Jposvmax[j].
\end{equation}

\subsection{Hyperparameter Tuning}
\label{app:hps}

\begin{table*}
    \vspace*{-1.7mm}  %
    \caption{Hyperparameters}
    \label{tab:hyperparameter}
    \centering
    \begin{tabular}{lCCcCCCC}  %
        \toprule
        & & & & \multicolumn{4}{c}{Best Value for each Optimization Scope} \\
         \cmidrule(lr){5-8}
        & \text{Hyperparameter} & \text{Search Space} & Sampling & \scopeM & \scopeMB & \scopeMQ & \scopeMBQ \\
        \midrule
        \multirow{7}{*}{\rotatebox[origin=c]{90}{\parbox{2cm}{\centering Gene specific}}} 
        & \nSimpleIK & 10 \leq \nSimpleIK \leq 1000 & log & \qty{135}{} & \qty{189}{} & - & - \\
        & \nCplxIK & \nSimpleIK \leq \nCplxIK \leq 1000 & log & 534 & 303 & - & - \\
        & \tplan & \realInt{0.1}{100} & log & \qty{3.68}{\second} & \qty{13.12}{\second} & \qty{2.07}{\second} & \qty{0.63}{\second} \\
        & \sigmaBase & \realInt{0}{1} & lin & - & \qty{0.19}{} & - & \qty{0.88}{} \\
        & \sigmaIK & \realInt{0}{1} & lin & - & - & \qty{0.13}{} & \qty{0.05}{} \\
        & \pLamarck & \realInt{0}{1} & lin & - & - & \qty{0.80}{} & \qty{1.00}{} \\
        & \nLamarck & 10 \leq \nLamarck \leq 1000 & log & - & - & \qty{148}{} & \qty{105}{} \\ \midrule
        \multirow{9}{*}{\rotatebox[origin=c]{90}{\parbox{3cm}{\centering All genes / Genetic alg.}}} & \dofInit & \min(4, \nGenes) \leq \dofInit \leq \min(8, \nGenes - 1) & lin & \qty{7}{} & \qty{7}{} & \qty{7}{} & \qty{8}{} \\
        & \chanceEmpty & \realInt{0}{0.8} & lin & \qty{0.50}{} & \qty{0.28}{} & \qty{0.58}{} & \qty{0.41}{} \\
        & \popSize & 5 \leq \popSize \leq 50 & lin & \qty{13}{} & \qty{17}{} & \qty{22}{} & \qty{35}{} \\
        & \nGenes & 3 \leq \nGenes \leq 21 & lin & \qty{13}{} & \qty{15}{} & \qty{15}{} & \qty{18}{} \\
        & \pMutate & \realInt{0.005}{0.3} & log & \qty{0.03}{} & \qty{0.06}{} & \qty{0.12}{} & \qty{0.08}{} \\
        & \nParentsMate & 1 \leq \nParentsMate \leq \popSize & lin & \qty{6}{} & \qty{12}{} & \qty{11}{} & \qty{9}{} \\
        & \nElites & \naturalInt{5} & lin & \qty{1}{} & \qty{3}{} & \qty{5}{} & \qty{2}{} \\
        & \nKeepParents & -1 \leq \nKeepParents \leq \nParentsMate& lin & \qty{3}{} & \qty{11}{} & \qty{11}{} & \qty{8}{} \\
        & \selectionPres & \realInt{1}{2} & lin & \qty{1.64}{} & \qty{1.74}{} & \qty{1.81}{} & \qty{1.17}{} \\
        \bottomrule
    \end{tabular}
\end{table*}

\begin{figure}[t]
    \vspace*{-1.7mm}  %
    \centering
    \begin{tikzpicture}
  \pgfplotstableread[col sep=comma]{figures/edf_data.csv}\edfdata  
  \begin{axis}[
      width=\linewidth,
      xlabel={Mean Cycle Time $\timeJ \left[ \unit{\second} \right]$},
      xmin=14,
      xmax=51,
      ylabel=Cummulative Probability,
      ymin=-0.03,
      ymax=1.02,
      legend style={fill opacity=0.8,
        draw opacity=1,
        text opacity=1,
        draw=none,
        font=\footnotesize,
        column sep=1pt,
        at={(axis cs:49.5,0.0)},
        anchor=south east},
      legend cell align={left},
      cycle list name=tumcolormark,
      line width=1pt
    ]
    \addlegendimage{empty legend}
    \addlegendentry{\hspace{-20pt}Optimization Scope}
    \addplot+ [mark=none] table[x=Objective,y=Modules_3600] {\edfdata};
    \addlegendentry{Modules $\assembly$}
    \addplot+ [mark=none, style=dashed] table[x=Objective,y=Modules+Base_xyz_3600] {\edfdata};
    \addlegendentry{$\assembly$ + Base $\B$}
    \addplot+ [mark=none, style=dashdotted] table[x=Objective,y=Modules+IK_3600] {\edfdata};
    \addlegendentry{$\assembly$ + IKs $\Jposs$}
    \addplot+ [mark=none,style=densely dotted] table[x=Objective,y=Modules+Base_xyz+IK_3600] {\edfdata};
    \addlegendentry{$\assembly$ + $\Jposs$ + $\B$}
  \end{axis}
\end{tikzpicture}  %
    \caption{Empirical cumulative density function of mean cycle time over 400 trials per optimization scope on the evaluation set containing four tasks from each task set.}
    \label{fig:hyperparam_cdf}
\end{figure}
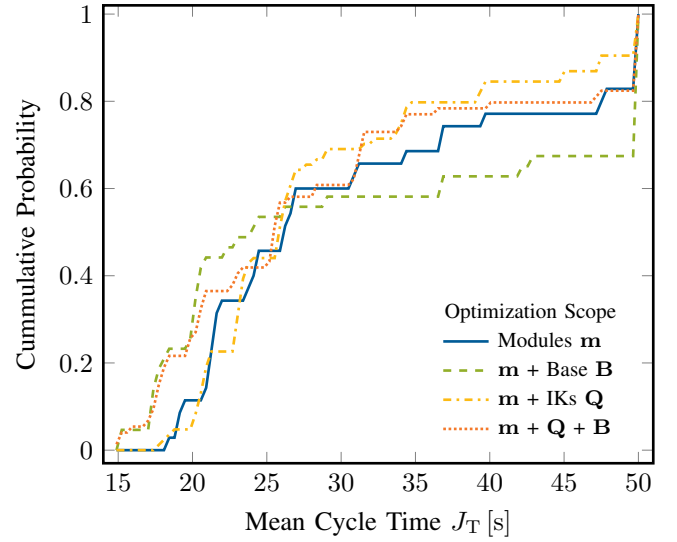

\noindent The considered search spaces and used hyperparameters per optimization scope are summarized in \cref{tab:hyperparameter}.
To judge the complexity and gains from hyperparameter tuning, we plot the cumulative density of different final costs, measured by the mean cycle time $\timeJ$ in \cref{fig:hyperparam_cdf}.
We observe that the scopes that optimize the base of the robot $\B$ show lower possible mean cycle time, which is consistent with scopes $\scopeMB$ and $\scopeMBQ$ solving significantly more tasks, especially in the hard task set (see \cref{fig:convergence}).
Overall, optimization of hyperparameters is beneficial as the mean tested hyperparameter results in roughly $\SI{5}{\second}$ to $\SI{10}{\second}$ slower mean cycle time.

We also tested whether the mean cycle time $\timeJ$ as the single optimization objective for hyperparameter tuning (including the fixed failure penalty for not found solutions $\costfail$) also optimizes the chance of success.
The Pearson correlation coefficient between the mean cycle time $\timeJ$ and the chance of solving a task is $\SI{-0.9993}{}$ with a p-value of $\SI{0.0000}{}$, i.e., there is almost a perfect negative correlation.
Finding hyperparameters that minimize the mean cycle time $\timeJ$, therefore, should also maximize the chance of success.

\subsection{Robot Modules}
\label{app:modules}

\begin{figure}
    \vspace*{-1.3mm}  %
    \centering
    \import{figures/}{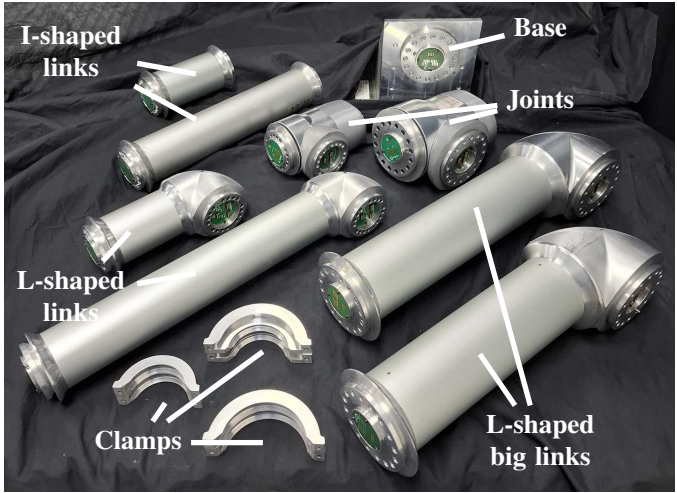}
    \caption{
    All module types inside modrob-gen2 manufactured by RobCo\cref{fn:robco} available in our lab.
    Modules come in two flange sizes: small (left) and big (right).
    All distal/output flanges point to the lower left of the image.
    }
    \label{fig:moduleSet}
\end{figure}

We optimize module compositions from the module set \textit{modrob-gen2}\footnote{Description: \censor{\href{https://cobra.cps.cit.tum.de/api/robots/modrob-gen2_5}{cobra.cps.cit.tum.de/api/robots/modrob-gen2\_5}}, \lastChecked} without limiting availability, i.e., $n_{\mathrm{avail}}$ in \cref{eq:cost:availableModules} is set to infinity for all modules.
The module set contains
\begin{itemize}
    \item two sizes (\textit{small} $\SI{86}{\milli\meter}$, \textit{big} $\SI{116}{\milli\meter}$) of revolute joints;
    \item two big L-shaped and nine small L-shaped static links;
    \item eight small I-shaped static links;
    \item four bases of both sizes and upward/ninety-degree rotated orientation.
\end{itemize}
A subset of modules available in our lab is shown in \cref{fig:moduleSet}.

\subsection{Additional Statistical Tests}
\label{app:addStatTest}

\noindent We provide statistical tests to show that the claimed differences in optimization scopes are significant at the end of the chosen optimization budget $\ttimeout = \SI{60}{\minute}$.
First, we provide the results of the Friedman tests in \cref{tab:friedman}, with the null-hypothesis being that there is no difference between the considered tests.
We conducted the test on the paired results of runs on the same task and seed.

Secondly, we use post-hoc tests to show which scopes significantly outperform one another based on Bonferroni-corrected p-values in \cref{tab:post_hoc_pairwise}.
Here, only the comparison of optimizing the assembly with the base ($\scopeMB$) or IK solutions ($\scopeMQ$) is \textit{not} significant for the chance of finding a valid solution.
This is consistent with \cref{fig:convergence}, where $\scopeMB$ is dominated by $\scopeMQ$ in the simple and hard sets of tasks, while the opposite holds for the real-world and edge-case sets of tasks.

\begin{table}
    \caption{Friedman test for difference in optimization scopes at $\ttimeout$}
    \label{tab:friedman}
    \centering
    \begin{tabular}{llll}
    \toprule
        Measure & Task set & Kendal's W & p-value \\
    \midrule
        Final Cost & All & $\SI{0.26}{}$ & $\SI{1.98e-122}{}$ \\
        Final Cost & Simple & $\SI{0.29}{}$ & $\SI{1.64e-66}{}$ \\
        Final Cost & Hard & $\SI{0.11}{}$ & $\SI{4.56e-7}{}$ \\
        Final Cost & Real-World & $\SI{0.20}{}$ & $\SI{1.01e-10}{}$ \\
        Final Cost & Edge & $\SI{0.36}{}$ & $\SI{9.75e-43}{}$ \\
    \midrule
        Solution Found & All & $\SI{0.10}{}$ & $\SI{1.48e-103}{}$ \\
        Solution Found & Simple & $\SI{0.08}{}$ & $\SI{1.56e-23}{}$ \\
        Solution Found & Hard & $\SI{0.12}{}$ & $\SI{5.24e-38}{}$ \\
        Solution Found & Real-World & $\SI{0.07}{}$ & $\SI{1.60e-5}{}$ \\
        Solution Found & Edge & $\SI{0.20}{}$ & $\SI{7.26e-62}{}$ \\
    \bottomrule
    \end{tabular}
\end{table}

\begin{table}
    \centering
    \caption{Pair-wise post-hoc Wilcoxon tests at $\ttimeout$}
    \label{tab:post_hoc_pairwise}
    \begin{tabular}{lllll}
    \toprule
         & Scope A & & Scope B & Corr. p-value \\
    \midrule
        \multirow{6}{*}{\rotatebox[origin=c]{90}{\parbox{2cm}{\centering Success Rate}}} & $\scopeM$ & $<$ & $\scopeMB$ & $\SI{1.61e-19}{}$ \\
        & $\scopeM$ & $<$ & $\scopeMQ$ & $\SI{7.19e-36}{}$ \\
        & $\scopeM$ & $<$ & $\scopeMBQ$ & $\SI{9.66e-85}{}$ \\
        & $\scopeMB$ & $<$ & $\scopeMQ$ & $\SI{1.43e-1}{}$ \\
        & $\scopeMB$ & $<$ & $\scopeMBQ$ & $\SI{2.22e-37}{}$ \\
        & $\scopeMQ$ & $<$ & $\scopeMBQ$ & $\SI{1.29e-28}{}$ \\
    \midrule
        \multirow{6}{*}{\rotatebox[origin=c]{90}{\parbox{2cm}{\centering Best Cycle Time $\timeJ$}}} & $\scopeM$ & $>$ & $\scopeMB$ & $\SI{8.52e-39}{}$ \\
        & $\scopeM$ & $>$ & $\scopeMQ$ & $\SI{1.21e-70}{}$ \\
        & $\scopeM$ & $>$ & $\scopeMBQ$ & $\SI{4.79e-88}{}$ \\
        & $\scopeMB$ & $>$ & $\scopeMQ$ & $\SI{8.42e-11}{}$ \\
        & $\scopeMB$ & $>$ & $\scopeMBQ$ & $\SI{5.46e-28}{}$ \\
        & $\scopeMQ$ & $>$ & $\scopeMBQ$ & $\SI{2.68e-4}{}$ \\
    \bottomrule
    \end{tabular}
\end{table}

\subsection{Sim-To-Real Gap}
\label{app:sim_real_gap}

\noindent We found the following differences between our numerical experiments and real-world validation:

\begin{itemize}
    \item There are slight variations in module mass and shape, as our modules are from various sub-versions; e.g., some newer joints have a longer housing due to altered break designs. 
    These variations might lead to unintended collisions or torques that are different than expected.
    \item The robot controller accepts only via points of the path, not a fully parameterized trajectory. 
    Therefore, the final execution time might not be equal to one parameterized by~\cite{Pham2018} implementing step~\ref{cost:traj_param} in \cref{ssec:hierElim}.
    Also, deviations between the planned and executed trajectory could lead to collisions.
    \item The physical joints might not be exactly calibrated, i.e., have a difference between the zero position in reality vs. the model.
    This may lead to deviations between the intended and executed path, leading to possible collisions with obstacles or missed goal poses.
    \item Differences between the scanned and real-world occupancy can result in collisions.
    \item We move the robot on an unmodelled sled (see \cref{fig:task_solutions}) in the plane of our lab table, i.e., the base pose is optimized in $\SETwo = \reals^2 \times \SOTwo$, parameterized by $\basev = \left[ x, y, \theta \right]^T$.
    \item Differences between the intended robot base pose $\B$ vs. the one set up in the real world can also lead to unintended collisions or missed goal poses.
    \item Unlike our setting for the simulation, we only have a limited number of modules in our lab (shown in \cref{fig:moduleSet}):
    \begin{itemize}
        \item two big joints (D116),
        \item four small joints (D86),
        \item big L-shaped links of length $\SI{350}{}, \SI{400}{\mm}$ (L116-350/400),
        \item small L-shaped links of length $\SI{165}{}, \SI{440}{\mm}$ (L86-165/440),
        \item small I-shaped links of length $\SI{150}{}, \SI{350}{\mm}$ (I86-150/350), and
        \item a single $\SI{90}{\degree}$ turned base.
    \end{itemize}
\end{itemize}

\subsection{Stress Test of Found Solutions}
\label{app:stressTest}

\noindent In addition to the validation experiments we also conducted simulated stress tests to analyze the robustness of the optimized robots and their trajectories.
Within the stress test, we varied the position and orientation of the goals within the two analyzed real-world tasks by up to $\SI{4}{\centi\meter}$ in random directions and $\SI{5}{\degree}$ in rotation, respectively.
In total, we defined $56$ altered goals per task resulting in $n = 560$ tests due to two tasks and five seeds each.

The previously found solutions were adapted to the altered tasks by adapting hierarchical elimination.
First, it was checked if the changed task was still fulfilled. 
If not, in a second step, we tried to find a new IK solution.
Third, if a new IK solution was found, we tested if it was enough to replace the last via point in the path with this new IK solution; if that was not enough, a new path planning was started.
Lastly, \cite{Pham2018} was applied for velocity planning.

Looking at the failure modes, we find that in $\SI{95}{\percent}$ ($n = 466 \; \text{of} \; 560$) of cases the chosen robot can be adapted just by reprogramming, i.e., finding a new solution trajectory.
This further supports our previous validation experiments, where the found solutions were adaptable to the real situation in $\SI{90}{\percent}$ of cases.
We find that $\SI{83.2}{\percent}$ ($n=466$) of altered tasks were still solved as the original solution did not stop at the edge of the goal tolerances.
Overall, the repairs failed in $\SI{2.3}{\percent}$ ($n=13$) of cases due to no valid IK being found.
In another $\SI{2.7}{\percent}$ ($n=15$) of cases, a valid path could not be found.
The remaining $\SI{11.8}{\percent}$ ($n=66$) were successfully repaired, resulting in a trajectory that completely fulfilled the altered task.

\ifcensor
\else
\subsection{Author Information}
\begin{IEEEbiography}[{\includegraphics[width=1in,height=1.25in,clip,keepaspectratio]{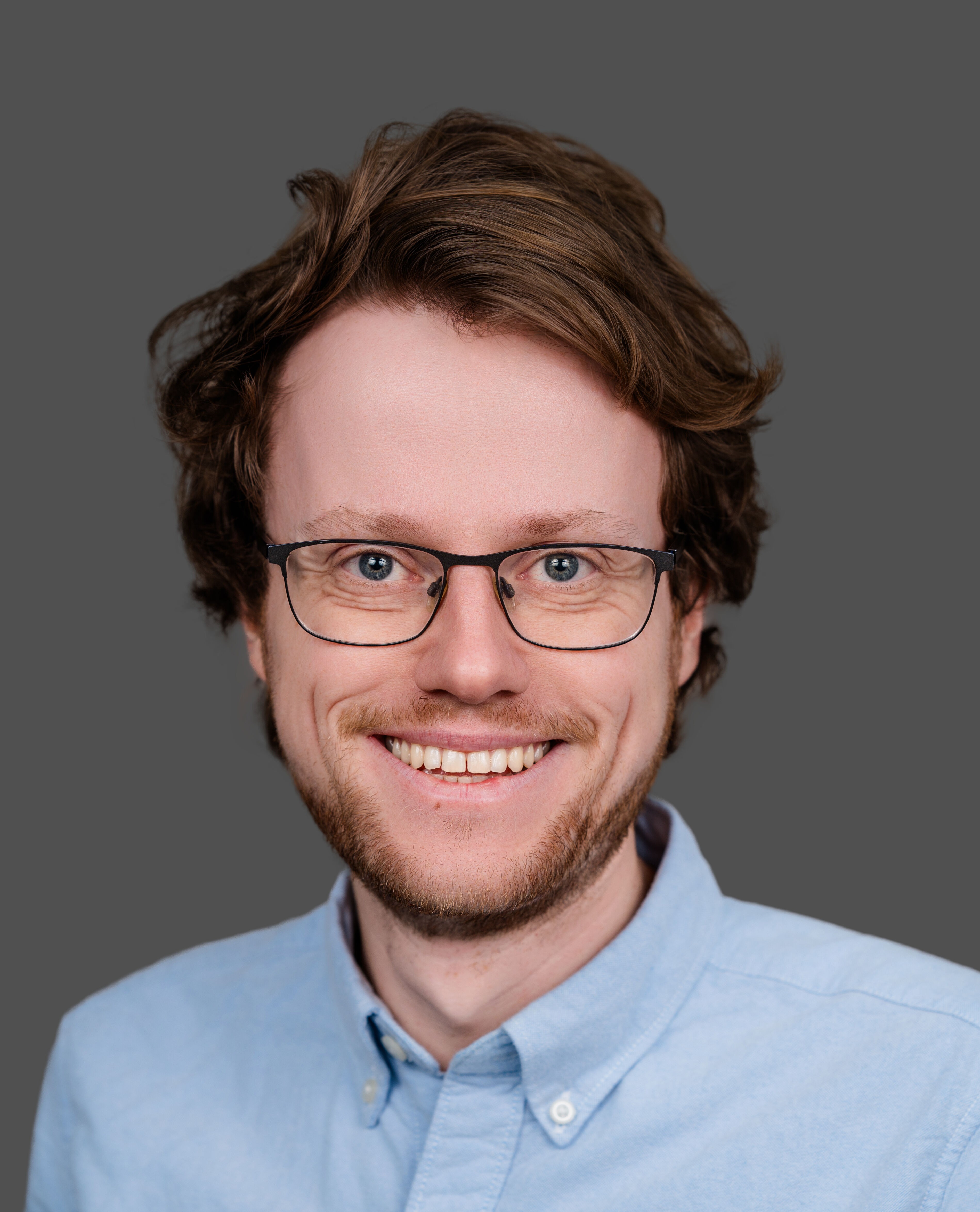}}]{Matthias Mayer}
is currently a Ph.D. candidate at the Technical University of Munich.
He received his B.Sc. degree in engineering sciences in 2017 and M.Sc. in robotics, cognition, and intelligence in 2019 from the same university.
He is interested in optimization-based robotic automation, including motion planning, co-designing robotic structures and behaviors, and benchmarking such systems.
\end{IEEEbiography}

\begin{IEEEbiography}[{\includegraphics[width=1in,height=1.25in,clip,keepaspectratio]{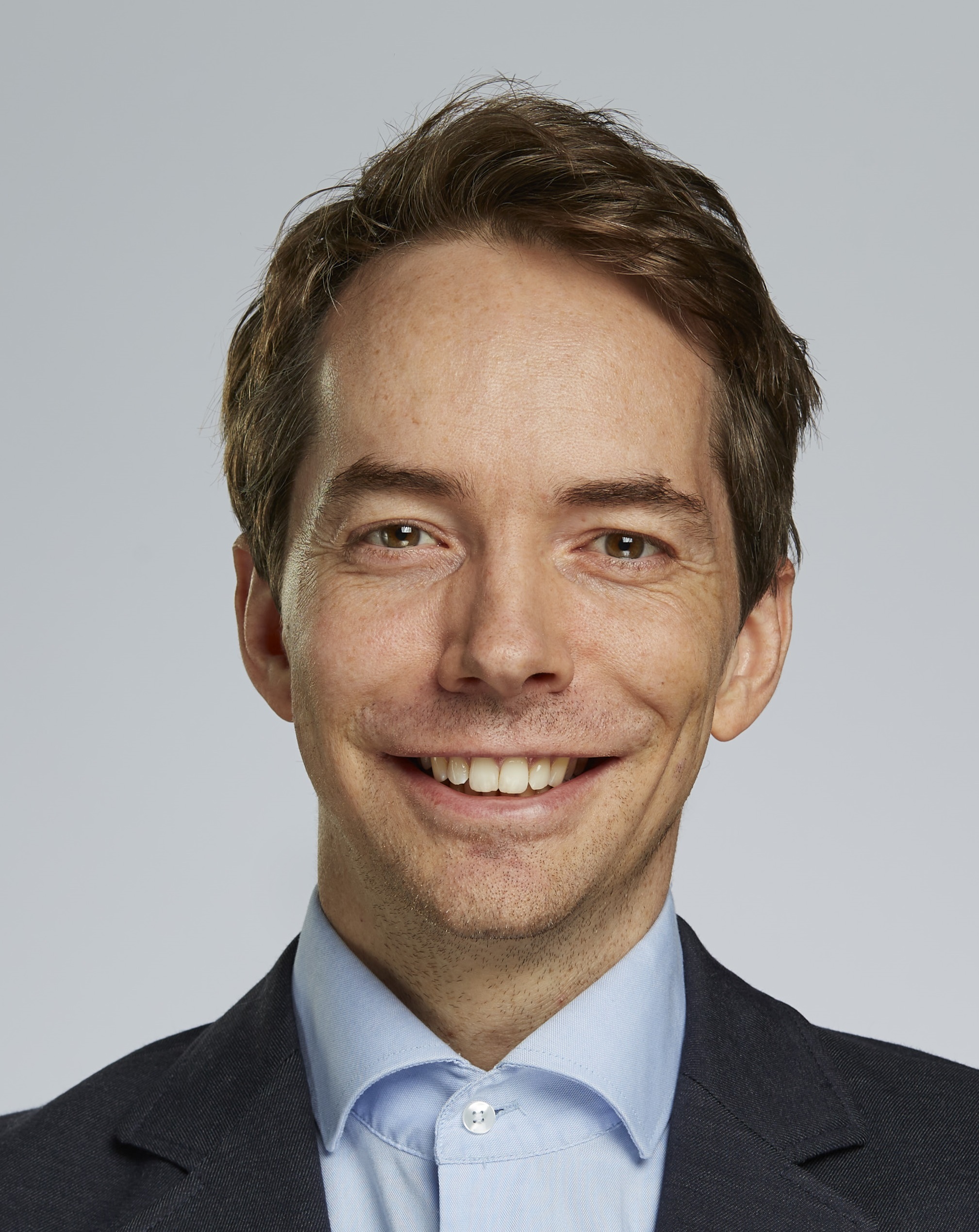}}]{Matthias Althoff}
received the Diploma Engineering degree in mechanical engineering and the Ph.D. degree in electrical engineering from the Technical University of Munich, Germany, in 2005 and 2010, respectively. 
He is currently an Associate Professor in computer science at the Technical University of Munich, Germany. 
From 2010 to 2012, he was a Postdoctoral Researcher with Carnegie Mellon University, Pittsburgh, PA, USA, and from 2012 to 2013, he was an Assistant Professor with Technische Universität Ilmenau, Germany. 
His research interests include formal verification of continuous and hybrid systems, reachability analysis, planning algorithms, nonlinear control, automated vehicles, power systems, and robotics.
\end{IEEEbiography}
\fi

\end{document}